\documentclass[10pt,twocolumn,letterpaper]{article}

\usepackage[pagenumbers]{iccv} % To force page numbers, e.g. for an arXiv version

\definecolor{iccvblue}{rgb}{0.21,0.49,0.74}
\usepackage[pagebackref,breaklinks,colorlinks,allcolors=iccvblue]{hyperref}

\def\paperID{9073} % *** Enter the Paper ID here
\def\confName{ICCV}
\def\confYear{2025}

\title{\emph{MoMa-Kitchen}: A 100K+ Benchmark for Affordance-Grounded \\Last-Mile Navigation in Mobile Manipulation}

\author{
Pingrui Zhang{$^{1, 2}$}\thanks{}
\quad Xianqiang Gao{$^{2, 3*}$}
\quad Yuhan Wu{$^{3}$}
\quad Kehui Liu{$^{2, 4}$} \\ 
\vspace{-8pt}\\ 
\quad Dong Wang{$^{2}$}
\quad Zhigang Wang{$^{2}$}
\quad Bin Zhao{$^{2, 4}$}
\quad Yan Ding{$^{2}$$^\dag$}
\quad Xuelong Li{$^{5}$} \\
\vspace{-8pt}\\
{\normalsize{$^{1}$}Fudan University \quad {$^{2}$}Shanghai AI Laboratory \quad {$^{3}$}University of Science and Technology of China\quad} \\{\normalsize
{$^{4}$}Northwestern Polytechnical University\quad
{$^{5}$}TeleAI, China Telecom Corp Ltd}\\
\small{\tt{\{zhangpingrui, dingyan\}@pjlab.org.cn}}\\ 
\vspace{-11mm}
}

\newcommand{\ours}{MoMa-Kitchen\xspace}
\newcommand{\ourmodel}{NavAff\xspace}

\begin{document}

\twocolumn[{%
\renewcommand\twocolumn[1][]{#1}%

\maketitle

\begin{center}
\centering
\captionsetup{type=figure}
\includegraphics[width=\textwidth]{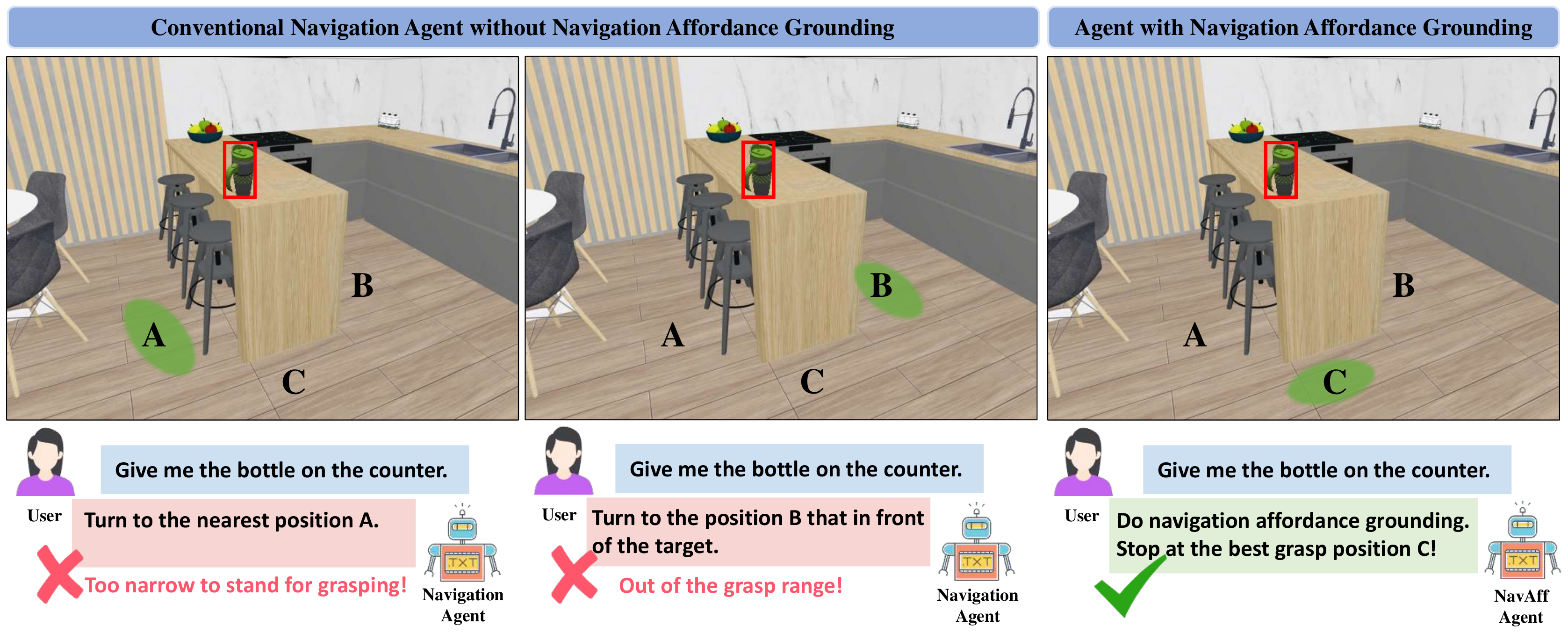}
\captionof{figure}
{
% \textbf{Comparison of navigation strategies for manipulation tasks.}  
Conventional navigation methods typically prioritize reaching a target location but do not account for constraints affecting manipulation feasibility.
\emph{Left}: Position \textbf{A} prioritizes proximity but is obstructed by chairs, preventing stable execution. 
\emph{Middle}: Position \textbf{B} places the robot in a spacious and stable area for operation but beyond its effective reach.  
\emph{Right}: Our approach, leveraging navigation affordance grounding, identifies Position \textbf{C} as the optimal stance, ensuring both reachability and task feasibility.
}
\label{fig:motivation}
\end{center}

}]

\begin{abstract}
In mobile manipulation, navigation and manipulation are often treated as separate problems, resulting in a significant gap between merely approaching an object and engaging with it effectively. Many navigation approaches primarily define success by proximity to the target, often overlooking the necessity for optimal positioning that facilitates subsequent manipulation. To address this, we introduce \ours, a benchmark dataset comprising over 100k samples that provide training data for models to learn optimal final navigation positions for seamless transition to manipulation. Our dataset includes affordance-grounded floor labels collected from diverse kitchen environments, in which robotic mobile manipulators of different models attempt to grasp target objects amidst clutter. Using a fully automated pipeline, we simulate diverse real-world scenarios and generate affordance labels for optimal manipulation positions. Visual data are collected from RGB-D inputs captured by a first-person view camera mounted on the robotic arm, ensuring consistency in viewpoint during data collection. We also develop a lightweight baseline model, \ourmodel, for navigation affordance grounding that demonstrates promising performance on the \ours benchmark. Our approach enables models to learn affordance-based final positioning that accommodates different arm types and platform heights, thereby paving the way for more robust and generalizable integration of navigation and manipulation in embodied AI. Project page: \href{https://momakitchen.github.io/}{https://momakitchen.github.io/}.

\end{abstract}    
\section{Introduction}
\label{sec:intro}

Most existing navigation algorithms define success \emph{in terms of} reaching a location near the target~\cite{shao2024moma}.
However, in household environments, navigation is an intermediate step preceding task-specific manipulation. 
As a result, such navigation strategies are inadequate for mobile manipulation tasks that requires precise end-effector positioning.
In practice, navigation and manipulation are tightly integrated.
For instance, when tasked with retrieving an object from a kitchen counter, a robot typically navigates toward the target using conventional policies before attempting manipulation.
While existing navigation algorithms reliably guide robots across rooms, they often fail in proximity to the target. 
The robot may stop beyond the manipulator’s reachable workspace or be obstructed by spatial constraints, rendering manipulation infeasible (as shown in~\cref{fig:motivation}).
Additionally, obstacles such as furniture, bins, or containers in cluttered environments are seldom accounted for in existing navigation policies, further complicating the selection of feasible grasping positions.
As a result, reliance solely on these navigation algorithms, without incorporating the demands of manipulation, limits their efficacy in addressing complex tasks in household settings.

This limitation highlights the disconnect between object localization (\emph{navigation}) and physical interaction (\emph{manipulation}) in mobile robotics. 
In particular, optimizing final positioning in the ``last mile'' remains a fundamental challenge, yet existing datasets and benchmarks provide limited supervision for this aspect. 
While recent efforts have employed large language models (LLMs) to assist in selecting optimal navigation positions, these approaches fall short when transitioning to the manipulation phase~\cite{zhu2024navi2gaze, zhi2024closedloop, qiu2024open}. 
Training-free LLMs struggle to accurately predict the requirements of robotic arm interactions. 
Additionally, they cannot dynamically adjust positioning strategies based on different robotic arm models or base morphology~\cite{NEURIPS2023_ee6630dc, zhou2024navgpt2}. 

To bridge this gap, we introduce \textbf{\ours}, a large-scale benchmark with over 100k samples designed to train models for affordance-grounded final positioning in mobile manipulation tasks. 
Our dataset comprises 127,343 episodes spanning 569 diverse kitchen scenes, where each episode involves predicting floor affordances that enable a robot to approach a target while avoiding collisions with obstacles. 
We collect large-scale floor affordance data by creating kitchen scenes with various layouts\textemdash in which target objects are either randomly placed or selected from common appliances (e.g., refrigerators and cabinets) and obstacles are strategically positioned to generate distinct scenarios. 
In the simulator, various mobile manipulators attempt to grasp target objects from multiple positions, and success rates are recorded to obtain ground truth affordance labels for each floor position. 
The collected visual data, consisting of RGB-D and point cloud inputs from a first-person view camera mounted on the robotic arm, along with robot-related information, are subsequently used to train our lightweight baseline model, \textbf{\ourmodel}, which identifies the optimal floor affordance region for subsequent manipulation. 
To enhance generalization, we employ various robotic arms (e.g., Flexiv and Franka) and varied mobile bases during grasping, enabling the model to learn that floor affordance predictions depend on arm height and operational range. 
This approach is intended to develop a model that can be used on heterogeneous devices~\cite{wang2024scaling}.
In summary, our contributions are as follows:
\begin{itemize}
    \item We propose \ours, the first large-scale dataset with over $100$k samples that bridges the gap between navigation and manipulation in mobile manipulation tasks by enabling models to optimize final positioning near target objects.
    \item We develop a fully automated data collection pipeline -- including scene generation, affordance labeling, and object placement -- to simulate diverse real-world scenarios and enhance model generalizability. 
    \item We design a lightweight baseline model, \ourmodel that employs RGB-D and point cloud inputs for navigation affordance grounding, achieving promising results on the \ours benchmark.
\end{itemize}
\section{Dataset Generation}
\label{sec:generation}

\textbf{Task Definition}.
\ours focuses on determining feasible final navigation positions that enable successful manipulation in cluttered environments. 
Given RGB-D inputs from a first-person camera and robot-specific parameters (e.g., arm reach, base height), the goal is to produce an affordance map over the floor. 
This map indicates where the robot can position itself to reliably manipulate the target while accounting for obstacles, bridging navigation and manipulation within a unified pipeline.

\begin{figure}[t!]
    \centering
    \includegraphics[width=\linewidth]{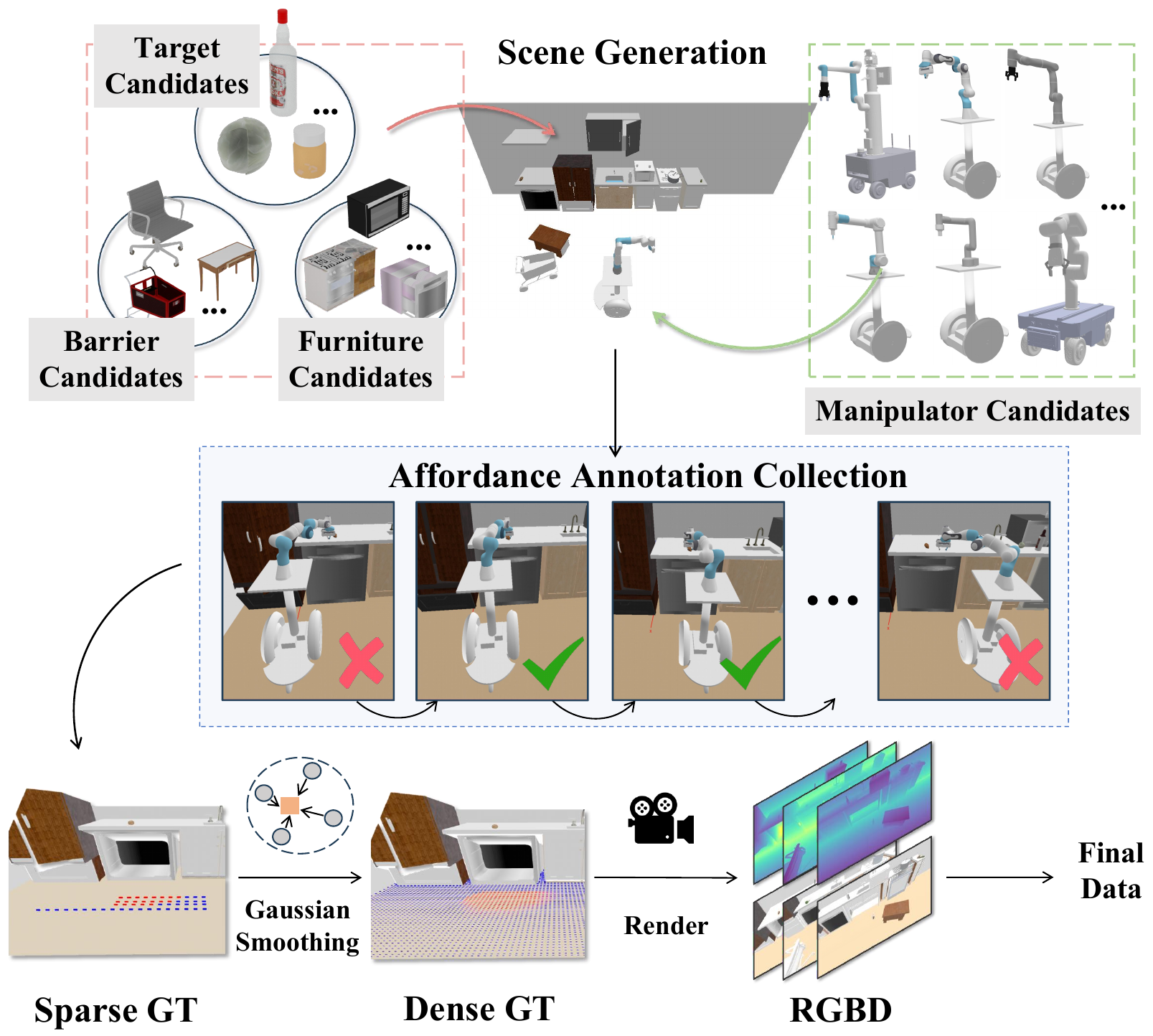}
    \caption{\textbf{Overview of scene setup and data generation pipeline.} Each scene features unique base furniture and layout, with randomly placed obstacles surrounding the target object to enhance scene complexity. Discrete navigation affordance values are collected by moving the mobile manipulator and interacting with the target objects in the scene. View transformation and Gaussian interpolation are then applied to generate a dense affordance map, along with corresponding RGBD data.}
    \label{fig:data_pipeline}
\end{figure}

\ours includes diverse kitchen environments with multiple types of robotic mobile manipulators, first-person view visual data, and navigation-specific affordance ground truth. 
As illustrated in~\cref{fig:data_pipeline}, we first construct large-scale kitchen environments that contain different targets, obstacles, and furniture. Mobile manipulators are then placed in these environments to obtain navigation affordance labels through robotic manipulation. 
For each labeled scene, we collect first-person view visual data from multiple randomly sampled robot viewpoints, including RGB-D images of the scene. The position of each viewpoint is also recorded to generate and transform both the global and floor point clouds. 
Additionally, floor-level affordance ground truth is collected near the target.

\subsection{Scene Setup}

To construct diverse kitchen environments for affordance annotation, we employ BestMan~\cite{Yang2024BestManAM}, a PyBullet-based simulation platform that integrates assets from PartNet-Mobility~\cite{Xiang_2020_SAPIEN, Mo_2019_CVPR, chang2015shapenet}. 
The generation process begins with a rectangular kitchen layout, where common kitchen furniture and appliances (e.g., sinks, cabinets, dishwashers, and fridges) are procedurally arranged along one wall. 
To ensure scene diversity, we randomize both the placement order of object categories and the specific instance selection within each category. 
We further augment each scene with both rigid and articulated objects as manipulation targets, and introduce additional obstacles around them to increase scene complexity.
% todo intro configurations
Once the scene assets are ready, we randomly select a manipulator from various types and place it into the scene to collect navigation affordance labels.

\subsection{Visual Data Collection}
\label{subsec:visual}

To comprehensively capture the target and its surroundings from the robot's perspective, we sample ten distinct camera poses $[\mathbf{R_c}|\mathbf{T_c}] \in \mathbb{R}^{4\times 4}$ around each target. 
These viewpoints are strategically selected to cover the target object, its surrounding environment, and the floor. For viewpoint selection, we alternate the positions to the left and right of the target object, gradually increasing the distance from it. After placing the robot at the selected position, we check whether the target object is within the observation range. If not, we randomly generate a new position. This process continues until the target object is within the line of sight, at which point we stop and record the viewpoint. 
For each camera pose, we collect RGB images $\mathbf{I}$, depth maps $\mathbf{D}$, and floor point clouds $ \mathbf{P}_\text{floor} \in \mathbb{R}^{n \times 3}$, where $n$ denotes the number of points in the point cloud.

\begin{figure*}[t]
    \centering
    \includegraphics[width=\linewidth]{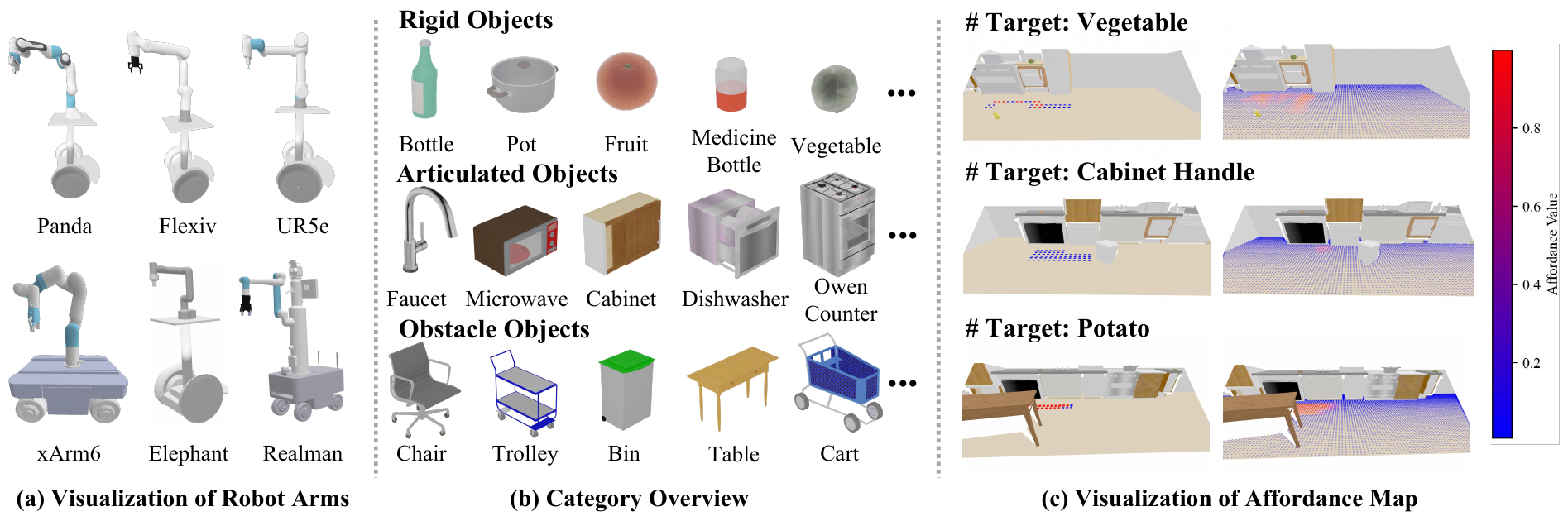}
    \caption{\textbf{(a) Robot arms used in \ours.} The end-effectors of the Panda, Flexiv, Elephant, Realman and xArm6 robots are grippers, while the end-effector of the UR5e is a suction cup. 
    \textbf{(b) Object categories utilized in \ours.} Each category consists of multiple object instances. 
    Rigid and articulated objects serve as manipulation targets, while obstacle objects are strategically placed around the target to enhance scene complexity. 
    \textbf{(c) Examples of affordance maps in \ours.} Discrete affordance values are first collected (left) by moving the mobile manipulator and allowing it to interact with the target. 
    Gaussian interpolation is then applied to obtain a smooth affordance map (right).}
    \label{fig:arms_category_affor}
\end{figure*}

\subsection{Affordance Labeling}
\label{subsec:affordance}

We gather mobile manipulators with various robotic arms to collect floor-level navigation affordance ground truth data for target objects. 
For each target object, we define the affordance sampling area $\mathcal{A}$ as a semicircular region on the floor, centered at the target's position with a radius equal to the maximum reach of the robotic arm. 
At each sampling position $\mathbf{p} \in \mathcal{A}$, we place the robot base and initialize the manipulator configuration, then align the end-effector orientation $\mathbf{R} \in \text{SO}(3)$ with the surface normal of either the target object (for rigid objects) or its functional link (for articulated objects). Here, $\text{SO}(3)$ denotes the Special Orthogonal Group, representing the set of all rotation matrices that describe rotations in three-dimensional space.

For each configuration $(\mathbf{p}, \mathbf{R})$, we attempt a manipulation and record a binary affordance outcome $v_p \in {0,1}$ at position $\mathbf{p}$. 
Success is determined by the outcome of the manipulation attempt: for robots equipped with two-finger parallel grippers (e.g., Panda, Flexiv, and xArm6), success is defined by a successful grasp of the target object; for the robot employing a suction-based end-effector (e.g., UR5e), success is determined by a valid suction-and-move action.
To associate these affordance values with the previously collected floor point cloud $\mathbf{P}_\text{floor}$, we first transform $v_p$ from the world coordinate system to the robot base frame. 
This transformation yields $v_{pbase}$ through:
\begin{equation}
\textstyle
     v_{pbase} = \mathbf{T}_{bc} \mathbf{T}_{cw} v_p, \label{eq:transform}
\end{equation}
where $\mathbf{T}_{bc}$ and $\mathbf{T}_{cw} \in \mathbb{R}^{4 \times 4}$ are the transformations from the camera to the base and from the world to the camera, respectively.
We then match each transformed affordance value $v_{pbase}$ to its nearest neighbor in $\mathbf{P}_\text{floor}$.
This association process can be formally expressed as:
\begin{equation}
    \textstyle
     \textbf{V}_{\text{aff}}(p_j) = 
            \begin{cases} 
            v_{pbase} & \text{if } \min_{p\in \mathcal{A}} \|p - p_j\| < \theta, \forall p_j \in \mathbf{P}_\text{floor},\\
            0     & \text{otherwise}, 
            \end{cases}
\end{equation}
where $\theta$ is a distance threshold. 
This yields sparse affordance values $\mathbf{V}_{\text{aff}} \in \mathbb{R}^{n \times 1}$.
To produce dense and continuous floor-level affordance maps, we employ Gaussian interpolation with k-nearest neighbors. 
This process produces interpolated affordance values $\hat{\mathbf{V}}_{\text{aff}} = \{\hat{v}_i \in [0, 1] \mid i = 1, \dots, n \}$ through the following formulation:
\begin{equation}
\textstyle
    \hat{v}_i = \frac{\sum_{j=1}^{k} w_{ij} v_j}{\sum_{j=1}^{k} w_{ij}}
\end{equation}
where the Gaussian weights $w_{ij}$ are computed as follows.
\begin{equation}
\textstyle
    w_{ij} = \exp\left(-\frac{\|p_i - p_j\|^2}{2\sigma^2}\right), \quad \forall i, j \label{eq:gaussian_interpolation}
\end{equation}
Here, $\|p_i - p_j\|$ is the Euclidean distance between the interpolation target $p_i$ and the sparse point $p_j$, $\sigma$ controls the width of the Gaussian kernel, and $s_j$ is the affordance value at $p_j$.
This weighting scheme ensures that points closer to $p_i$ have a larger influence, resulting in a smooth, continuous affordance map that captures the spatial distribution of potential manipulation outcomes.

\subsection{Generated Dataset Statistics}

As summarized in Table~\ref{tab:data_split}, \ours spans 569 kitchen scenes, each with procedural variations in objects and obstacles. These configurations yield over 127k episodes, each capturing an RGB-D view and the resulting affordance map. 
Six robot arms (Flexiv, Panda, UR5e, xArm6, Realman, Elephant) are deployed to ensure diverse reachability constraints and end-effector types. (see~\cref{fig:arms_category_affor}(a))
The dataset covers 137 kitchen-relevant assets (65 rigid, 48 articulated, and 24 obstacle types), providing a broad spectrum of layouts and manipulation targets. (see~\cref{fig:arms_category_affor}(b)) (Details in Supp.1)
During data collection, each configuration is labeled via discrete success/failure outcomes for different robot placements. 
These discrete labels are then interpolated to produce dense affordance maps (see~\cref{fig:arms_category_affor}(c)). 
Collectively, \ours provides large-scale supervision for end-to-end training of navigation-to-manipulation models that generalize across varying hardware and scene complexity.

% \subsection{Dataset Statistics}
\begin{table}[t]
    \footnotesize
    \centering
    \caption{\textbf{Dataset Split Statistics.}}
    \resizebox{1.0\linewidth}{!}{
        \begin{tabular}{@{}lccc@{}}
            \toprule
            Split                & Scenes & Configurations & Episodes\\
            \midrule
            Train                & $456$ & $11,408$ & $102,687$ \\
            Test (Unseen Scenes) & $113$ & $2,747$ & $24,656$ \\
            \bottomrule
            \end{tabular}
    }
    
    \vspace{5pt}
    
    \label{tab:data_split}
    \vspace{-10pt}
\end{table}

\section{Baseline Model}
\label{sec:baseline}

\begin{figure*}[t]
  \centering
  % \fbox{\rule{0pt}{2in} \rule{1.0\linewidth}{0pt}}
   \includegraphics[width=1.0\textwidth]{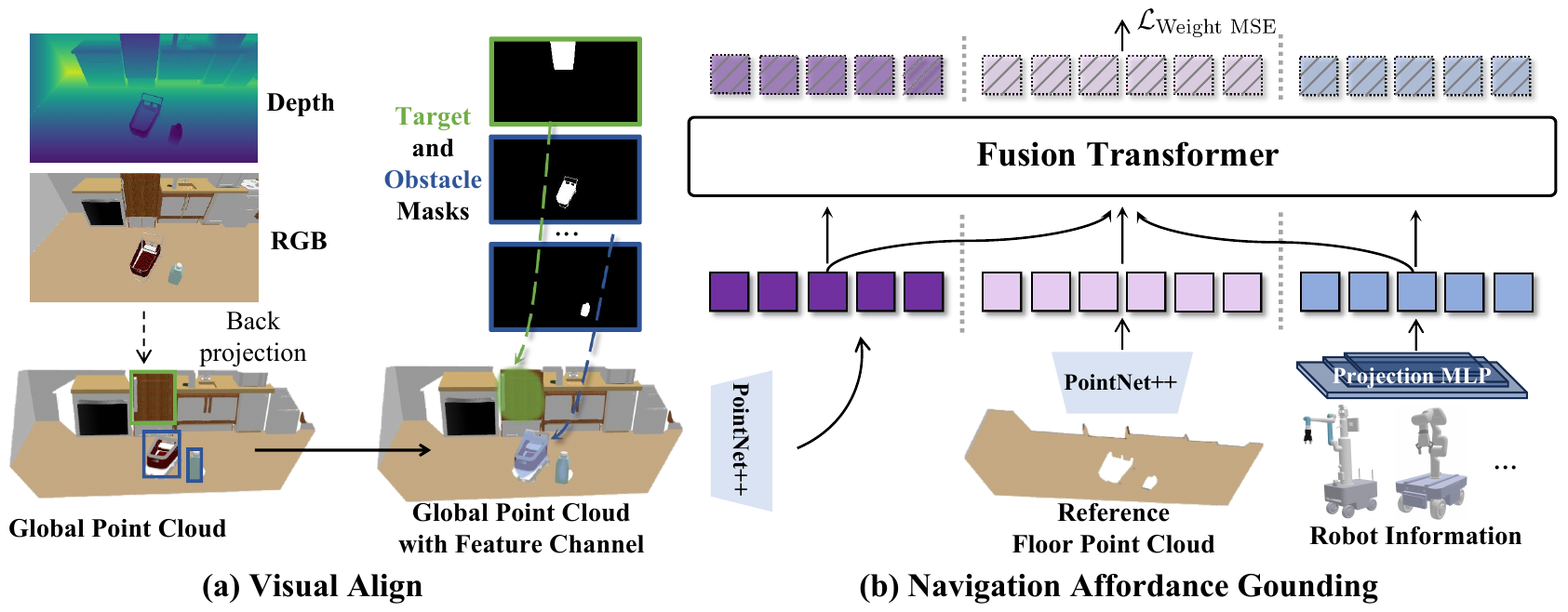}

   \caption{\textbf{NavAff Baseline.} \textbf{(a) Visual Alignment Module:} Projects object masks to align 2D visual features with 3D spatial representations. \textbf{(b) Navigation Affordance Grounding Module:} Fuses global point cloud, floor point cloud, and robot-specific features to predict navigation affordance maps.}
   \label{fig:baseline}
    % \vspace{-5pt}
\end{figure*}

In this section, we introduce our baseline model, \ourmodel, which is designed for optimal manipulation positioning and obstacle interaction in complex mobile manipulation tasks. 
% \subsection{Overview}
As shown in ~\cref{fig:baseline}, Our baseline method consists of two main components: Visual Alignment Module (\textbf{VAM}) and Navigation Affordance Grounding Module (\textbf{NAG}). This two-part structure is designed to process visual and spatial information sequentially, facilitating effective affordance prediction for mobile manipulation tasks.

\subsection{Visual Alignment}

VAM extracts and projects features of target object $\mathbf{T}$ and obstacles $\{\mathbf{O}_1, \mathbf{O}_2,\dots, \mathbf{O}_n\}$ onto the global point cloud $\mathbf{P}_{global} \in \mathbb{R}^{3 \times N_{global}}$, to ensure accurate 3D representation of them.
As illustrated in ~\cref{fig:baseline}(a), VAM processes an image $\mathbf{I} \in \mathbb{R}^{3 \times H \times W}$ captured by the robot’s camera along with its corresponding depth $\mathbf{D} \in \mathbb{R}^{H \times W}$, extracting target and obstacle masks using BestMan\cite{Yang2024BestManAM}, where $H, W$ denotes the height and width of the camera observation. 
The global point cloud $\mathbf{P}_{global}$ is then generated via inverse projection. 
To enhance feature representation, we leverage PointNet++~\cite{qi2017pointnet++}, which supports additional feature channels in the point cloud input.
Pixels from the target mask are projected into the 3D space to form a \textit{target channel}, with values at target locations set to $1$, while obstacle mask pixels create an \textit{obstacle channel}, assigned related values of -1~\cite{Huang2023VoxPoserC3}.
This process enriches the global point cloud by embedding 2D visual cues (e.g., target and obstacle features) into the 3D spatial structure. 
The resulting feature-enhanced point cloud $\mathbf{\Bar{P}}_{global}$ serves as input for the next stage, aligning visual perception with spatial information for improved downstream processing.

\subsection{Navigation Affordance Grounding}

NAG uses the robot's relevant information and the visually aligned global point cloud to interact with the reference floor, generating a robot-specific affordance prediction for optimal positioning. As illustrated in~\cref{fig:baseline}(b), NAG takes as input the feature-enhanced global point cloud $\mathbf{\Bar{P}}_{global}$ from VAM, the floor point cloud $\mathbf{P}_{floor}$, and the robot-related information $\mathbf{R}_{I}$. 
To extract features from the point cloud data,  PointNet++ processes $\mathbf{\Bar{P}}_{global}$ and $\mathbf{P}_{floor}$, yielding feature representations $\mathbf{F}_{pcg}$ and $\mathbf{F}_{pcf}$, respectively.
% For the point cloud data, PointNet++ extracts features from $\mathbf{\Bar{P}}_{global}$ and $\mathbf{P}_{floor}$, resulting in $\mathbf{F}_{pcg}$ and $\mathbf{F}_{pcf}$, respectively.
These features are then tokenized using a multi-layer perceptron (MLP) in preparation for multi-head cross-attention~\cite{vaswani2017attention}: 
\begin{align}
     \mathbf{\bar{F}}_{pcg} &= \operatorname{Tokenizer}(\mathbf{F}_{pcg}),\\
     \mathbf{\bar{F}}_{pcf} &= \operatorname{Tokenizer}(\mathbf{F}_{pcf}),
\end{align}
where $\mathbf{\bar{F}}_{pcg} \in \mathbb{R}^{n_{global} \times dim}$ and $\mathbf{\bar{F}}_{pcf} \in \mathbb{R}^{n_{floor} \times dim}$. 

For the mobile manipulator, $\mathbf{R}_{I}$ encodes the base platform height and the operational radius of the robotic arm. 
This information is also processed by an MLP to produce the robot-specific tokens $\mathbf{F}_{r}$, which are then concatenated with $\mathbf{\bar{F}}_{pcg}$, forming a combined key and value for cross-attention fusion with $\mathbf{\bar{F}}_{pcf}$. The process is defined as follows:
\begin{equation}
    \mathbf{F}_{r} = \operatorname{Tokenizer}(\operatorname{MLP}(\mathbf{R}_{I}))
\end{equation}
\begin{equation}
    \mathbf{F}_{out} = \operatorname{MCA}(\mathbf{\bar{F}}_{pcf}\mathbf{W}_{q},[\mathbf{\bar{F}}_{pcg},\mathbf{F}_{r}]\mathbf{W}_{k},[\mathbf{\bar{F}}_{pcg},\mathbf{F}_{r}]\mathbf{W}_{v})
\end{equation}
where $\mathbf{F}_{r} \in \mathbb{R}^{n_{robot} \times dim}$, $\mathbf{F}_{out} \in \mathbb{R}^{n_{floor} \times dim}$. 
Finally, $\mathbf{F}_{out}$ is passed through a decoder $f_d$ to predict the navigation affordance grounding $\mathbf{P}_{out}$.  

\subsection{Loss Function}
Following standard practices in affordance grounding, we utilize the mean squared error (MSE) loss as the primary objective function to align the model's predictions $\mathbf{P}_{out}$ with the ground truth affordance labels $\mathbf{P}_{label}$. In our experiments, we observed that zero-valued elements constitute a large proportion of the ground truth floor affordance labels, leading to an imbalance between zero-valued elements and those with non-zero values. To address this issue, we apply a weighted mask to perform a weighted average on the MSE loss (Weighted MSE).
The Weighted MSE loss is computed by adjusting the weight for zero-valued elements in the ground truth. Specifically, for elements in the ground truth that are zero, we apply a weight of \(\lambda\ \) with a probability of 0.5, while other elements are assigned a weight of 1. This ensures the model places adequate emphasis on non-zero affordance areas.
The formula for the Weighted MSE loss is as follows:
\begin{align}
\mathcal{L}_{\text{Weight MSE}} = \frac{1}{N} \sum_{i=1}^{N} \mathbf{W}_i \cdot (\mathbf{P}_{out, i} - \mathbf{P}_{label, i})^2
\end{align}
where
\begin{align}
\mathbf{W}_i = 
\begin{cases} 
\lambda & \text{if } \mathbf{P}_{label, i} = 0 \text{ and with prob. 0.5} \\
1 & \text{otherwise.}
\end{cases}
\end{align}
Here, \(N\) represents the total number of elements, \(\mathbf{W}_i\) is the weight applied to each element based on the corresponding floor ground truth, and \(\lambda\ \in(0, 1) \) is a hyperparameter that assigns a reduced weight to samples with a zero ground truth value, applied with a probability of 0.5.

\section{Experiments}
\label{sec:experiment}

In this section, we benchmark \ourmodel on the \ours dataset using the data split described in \cref{tab:data_split}, with affordance annotations from \cref{subsec:affordance} and visual data from \cref{subsec:visual}. We introduce baseline results for \ourmodel, conduct a comprehensive performance analysis across both simulated and real-world environments, and identify emerging challenges in complex mobile manipulation scenarios that are uniquely highlighted through our benchmark. 

\begin{table}[htbp]
    \centering
    \caption{\textbf{Main results.} Quantitative evaluation of navigation affordance grounding performance of \ourmodel.}
    \resizebox{1.0\linewidth}{!}{
        \begin{tabular}{@{}llcccc@{}}
            \toprule
            & Method & \textbf{RMSE} $\downarrow$ & \textbf{logMSE} $\downarrow$ & \textbf{PCC} $\uparrow$ & \textbf{SIM} $\uparrow$  \\
            \midrule
            & PointNet++~\cite{qi2017pointnet++} & $0.164$ & $0.0142$ & $0.565$ & $0.589$  \\
            & VoteNet~\cite{Qi_2019_ICCV} & $0.167$ & $0.0143$ & $0.543$ & $0.570$  \\
            & H3DNet~\cite{Zhang2020H3DNet3O} & $0.174$ & $0.0156$ & $0.503$ & $0.522$  \\
            & NavAff & $\textbf{0.147}$ & $\textbf{0.0115}$ & $\textbf{0.680}$ & $\textbf{0.696}$  \\
            \bottomrule
            \end{tabular}
    }
    % \vspace{5pt}
    \label{tab:main_result}
    \vspace{-8pt}
\end{table}

\subsection{Experimental Settings}
Given that this is a newly proposed task, no existing methods provide a direct basis for comparison. To establish an evaluation, we adapt three well-established models from point cloud learning: PointNet++\cite{qi2017pointnet++}, VoteNet~\cite{Qi_2019_ICCV}, and H3DNet~\cite{Zhang2020H3DNet3O}. Specifically, PointNet++ serves as a foundational backbone model widely used in point cloud processing tasks; VoteNet and H3DNet, both originally designed for 3D object detection, are suited for adaptation to navigation affordance grounding on our benchmark. By leveraging these models, we aim to establish strong baselines and gain insights into how existing point cloud techniques perform when applied to this new challenge. 

For evaluation, we employ standard and diversity metrics tailored for the \ours benchmark. Each metric provides a distinct perspective on navigation affordance grounding performance: Root Mean Squared Error (\textbf{RMSE}), facilitating understanding the magnitude of prediction error; Logarithmic Mean Squared Error (\textbf{logMSE}) focuses more on relative differences rather than absolute differences; Pearson Correlation Coefficient (\textbf{PCC}) helping to gauge the model's ability to maintain consistent patterns with the ground truth data across various affordance regions; Cosine Similarity (\textbf{SIM}) compares the structural or shape similarity between predicted and ground truth.

All the above experiments are trained on a single NVIDIA A100 GPU with a batch size of 64, using the Adam optimizer with a learning rate of 8e-4. Further experimental details are available in the Appendix.

\subsection{Quantitative Analysis}

\subsubsection{Main Results}
\cref{tab:main_result} reports the metrics of the proposed method \textbf{\ourmodel} compared with other methods on the \textbf{\ours} benchmark. The results highlight that \ourmodel achieves superior performance in navigation affordance grounding, outperforming transferred methods across all evaluated metrics. Specifically, \ourmodel achieves an RMSE of 0.147 and a logMSE of 0.0115, demonstrating a marked improvement in prediction accuracy compared to the second-best method, PointNet++. In terms of correlation and similarity measures, \ourmodel achieves the highest scores with a PCC of 0.680 and SIM of 0.696, highlighting its robustness in capturing affordance patterns even in cluttered environments. Notably, VoteNet and H3DNet fall short of \ourmodel's accuracy and consistency across the various metrics, showcasing the effectiveness of our approach in the proposed \ours benchmark. 

For further analysis, while VoteNet and H3DNet are effective in point cloud detection and proposal classification tasks, they are less suited to the fine-grained requirements of navigation affordance grounding, resulting in slightly lower performance. All methods are trained for the same number of epochs, but H3DNet converges more slowly due to its larger parameter count. Consequently, within the same training duration, H3DNet demonstrates the weakest performance. These findings suggest that a lightweight model capable of fine-grained feature prediction may achieve superior results on our \ours benchmark. 

\begin{figure*}[t]
    \vspace{-5pt}
    \centering
    \includegraphics[width=\linewidth]{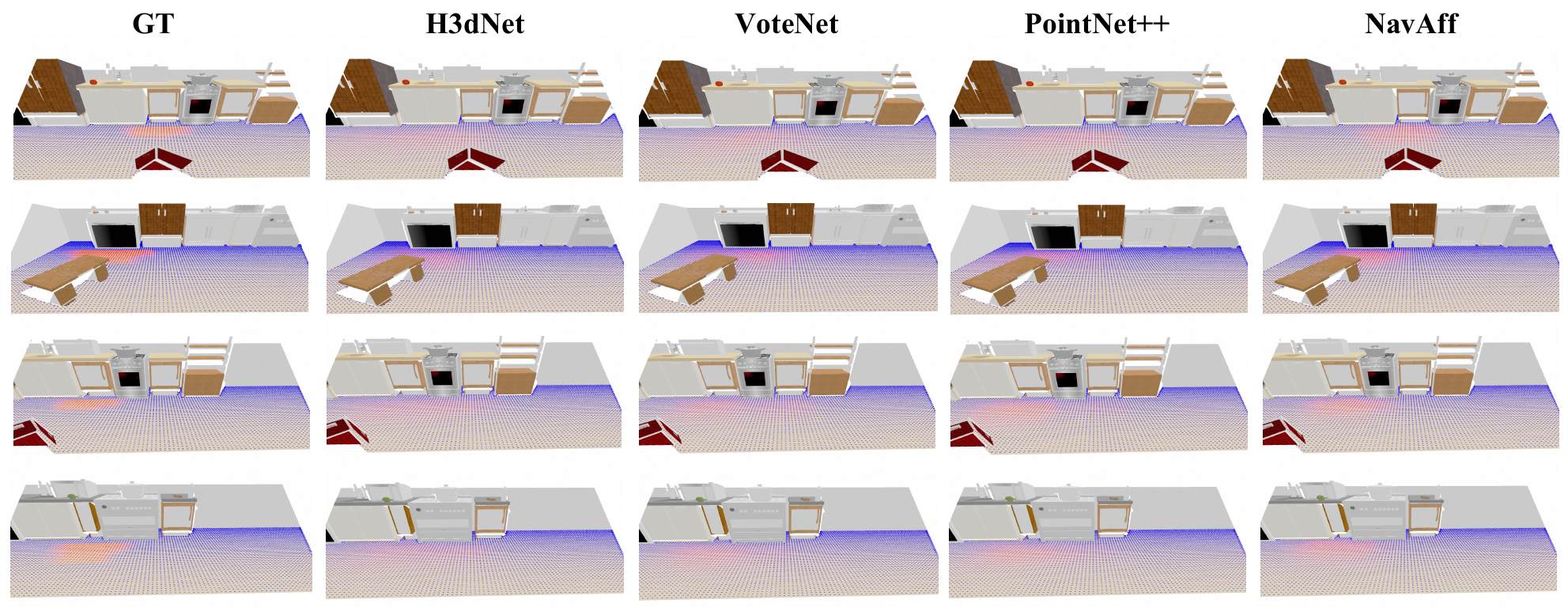}
    \caption{\textbf{Qualitative comparison of navigation affordance between all methods and ground truth.} Blue to red regions indicate affordance values ranging from $0$ to $1$, while void areas represent obstacle-occupied spaces.}
    \label{fig:exp_visualization}
    \vspace{-10pt}
\end{figure*}

\subsubsection{Manipulation Success Rate}

\begin{table}[t]
    \centering
    
    \caption{\textbf{Manipulation Success Rate (MSR)}.  }
    \vspace{-2pt}
    \resizebox{1.0\linewidth}{!}{
        \begin{tabular}{@{}lccccc@{}}
            \toprule
            \textbf{MSR}                & Random & H3DNet & VoteNet & PointNet++ & NavAff\\
            \midrule
            \texttt{Top1} & 0.080 & 0.54 & 0.56 & 0.60 & \textbf{0.72}\\
            \texttt{Top5} & 0.046 & 0.47 & 0.53 & 0.58 & \textbf{0.66}\\
            \bottomrule
            \end{tabular}
    }
    
    \label{tab:SR_evaluation}
    \vspace{-10pt}
\end{table}

To more intuitively demonstrate the direct improvement of our method and benchmark its performance in mobile manipulation, we introduce the manipulation success rate (\textbf{MSR}). The results are shown in ~\cref{tab:SR_evaluation}. The \texttt{Top1} \textbf{MSR} refers to moving the robot to the location with the highest affordance score and recording the MSR in the test scenes. The \texttt{Top5} \textbf{MSR} refers to recording the average MSR across the top 5 locations.

As quantitatively shown in ~\cref{tab:SR_evaluation} using the \textbf{MSR} metric, our navigation affordance prediction paradigm shows significant practical advantages. A comparison between the ``Random'' baseline and other methods reveals that approaches trained on \ours benchmark significantly outperform those relying on randomly sampled points within the robot's operational radius, thereby validating the effectiveness of our benchmark. Among all the trained methods, \ourmodel sets a new state-of-the-art performance, achieving an impressive \texttt{Top1} \textbf{MSR} of 72\% and \texttt{Top5} \textbf{MSR} of 66\%. Notably, when expanding the candidate point selection from Top1 to Top5, the model demonstrates strong generalization with minimal performance degradation, suggesting its robust adaptability to varying operational conditions. 

Additionally, by comparing the MSR values of all models, we observe a strong positive correlation between MSR and the accuracy of navigation affordance prediction (see~\cref{tab:main_result} and ~\cref{tab:SR_evaluation}), indicating that accurate navigation affordance prediction indeed contributes to higher manipulation success rates in mobile manipulation tasks.

\subsection{Qualitative Results}

We show the qualitative results of \ourmodel and compared methods for navigation affordance grounding in~\cref{fig:exp_visualization}. As illustrated, our model successfully predicts floor-level affordance maps that closely align with the ground truth patterns. The visualization clearly reveals distinct void regions in both predicted and ground truth maps, which correspond to areas occupied by obstacles, effectively demonstrating our model's ability to recognize spatial constraints. Compared to other methods, \ourmodel exhibits a more accurate representation of these void regions, indicating its superior capability in handling complex spatial relationships and navigating cluttered environments.

\subsection{Ablation Study}

To understand the impact of each module in our proposed \ourmodel model, we conduct an ablation study by progressively removing components and evaluating performance on the \ours benchmark. \cref{tab:ablation} shows the results of this study, where we assess the model variations by excluding robot information, the Visual Alignment Module (VAM), and the global point cloud. 
Below, we analyze the role and impact of each component in detail.

\begin{itemize}
    \item \textbf{\emph{w/o} robot information.} Removing robot-specific parameters (\eg, base height and arm reach) leads to moderate performance degradation, particularly in RMSE and logMSE. The limited impact may be attributed to the simplified representation of robot information in \ourmodel, which only considers two parameters. For instance, only xArm6 has a unique base height, while other platforms share identical base configurations. Future work could explore incorporating additional robot-specific parameters to enhance performance.

    \item \textbf{\emph{w/o} VAM.} Excluding the Visual Alignment Module significantly reduces performance, particularly in PCC and SIM metrics, underscoring the importance of integrating 2D visual cues for accurate navigation affordance prediction. This demonstrates that aligning visual data with spatial information is essential for accurate and robust navigation affordance grounding.

    \item \textbf{\emph{w/o} global point cloud.} Without the global point cloud, the model's performance drops across all metrics, with the largest degradation in PCC. This indicates that global spatial context is crucial for capturing the layout of objects and obstacles in the scene.
\end{itemize}

\begin{table}[t]
    \centering
    \caption{\textbf{Ablation study results of \ourmodel.} }
    \vspace{-2pt}
    \resizebox{1.0\linewidth}{!}{
        \begin{tabular}{ll|ccccc}
            \toprule
            &  & \textbf{RMSE} $\downarrow$ & \textbf{logMSE} $\downarrow$ & \textbf{PCC} $\uparrow$ & \textbf{SIM} $\uparrow$  & \texttt{top1}\textbf{MSR} $\uparrow$\\
            \midrule
            & NavAff & $0.147$ & $0.0115$ & $0.680$ & $0.696$ & $0.72$ \\
            &  \emph{w/o} robot information & $0.148$ & $0.0115$ & $0.670$ & $0.688$ & $0.70$ \\
            &  \emph{w/o} VAM & $0.165$ & $0.0140$ & $0.562$ & $0.589$ & $0.63$ \\        
            &  \emph{w/o} global point cloud & $0.168$ & $0.0144$ & $0.534$ & $0.568$ & $0.58$ \\
            \bottomrule
            \end{tabular}
    }
    % \vspace{5pt}
    \label{tab:ablation}
    \vspace{-8pt}
\end{table}

\begin{figure}[htbp]
    \centering
    \includegraphics[width=\linewidth]{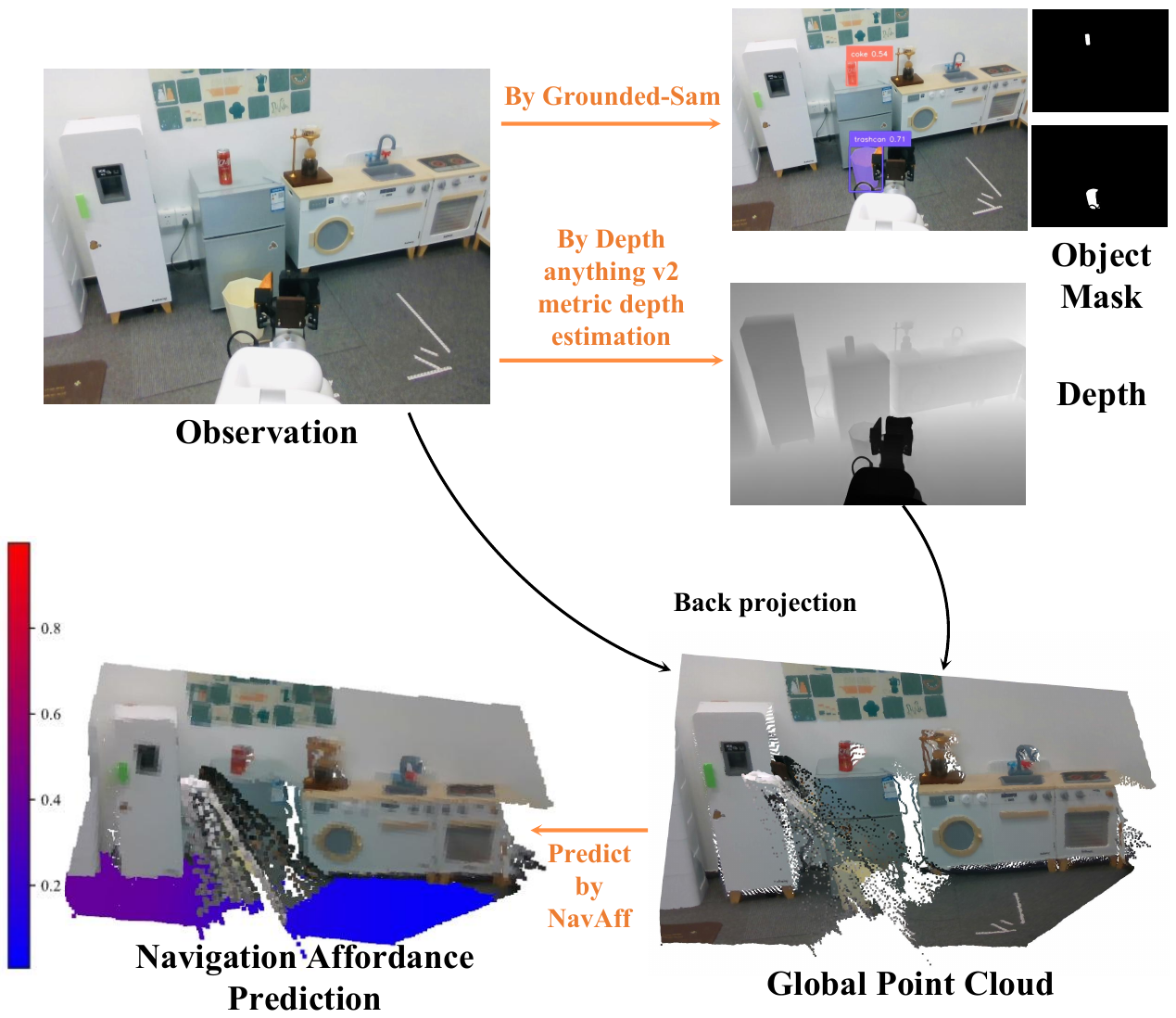}
    \caption{\textbf{Pipeline of real-world application with simulator trained \ourmodel.} The D435i camera mounted on the mobile manipulator captures image, and our model predicts optimal navigation affordances on the floor based on this data.}
    \label{fig:real_world_application}
    \vspace{-12pt}
\end{figure}

\subsection{Real World Application}

As shown in ~\cref{fig:real_world_application}, we validate our method by conducting experiments in a real-world setting. 
A D435i camera is mounted on the head of a mobile manipulator to capture data. Using Grounded-Sam~\cite{Ren2024GroundedSA}, we obtain open-vocabulary object masks, while Depth Anything v2~\cite{Yang2024DepthAV} provided the depth images. We generate the global point cloud through back projection and segmented the object’s local point cloud based on the mask. Then, we apply our model—trained on simulation data—to predict navigation affordances on the floor. 

The prediction results, shown in ~\cref{fig:real_world_application}, demonstrate that our model performs well in real-world kitchen scenarios, validating the strong generalization capability of our benchmark. This success highlights the robustness of our method, which was specifically designed to minimize the visual gap between simulation and reality. By addressing this challenge, our model is able to effectively adapt to real-world settings, showing that our approach is not only effective in controlled environments but also highly reliable in dynamic, real-world situations. Details of mobile manipulation demo are available in the supplementary material.

\section{Related Work}
\label{sec:related}

% \subsection{Mobile Manipulation}
\noindent\textbf{Mobile Manipulation}.
Such tasks have been extensively studied by researchers in both simulated and real-world settings~\cite{fu2023deep,fu2024mobile,mittal2022articulated,srivastava2022behavior,wong2022error,wu2023m,yokoyama2023asc}. 
More recently, approaches that integrate visual and linguistic information have emerged as promising methods for achieving unified reasoning in mobile manipulation~\cite{yenamandra2023homerobot,liu2024ok,qiu2024open,zhi2024closed,wang2024mosaic}. 
However, most existing methods still lack the versatility required to seamlessly combine coarse and fine motions for both navigation and manipulation. 
Prior mobile manipulation approaches often assume obstacle-free or easily navigable environments.
While interactive navigation techniques~\cite{xia2020interactive} attempt to tackle scenarios in which obstacles must be moved or manipulated\textemdash such as shifting boxes or pressing buttons\textemdash they typically rely solely on geometric reasoning~\cite{stilman2005navigation,van2008interactive,wang2020affordance,li2020hrl4in,zeng2021pushing,wang2024camp}. 
In contrast, our proposed benchmark is specifically designed to address these limitations by integrating both semantic and geometric reasoning for optimal positioning and manipulation.
This benchmark provides a robust and cohesive foundation for advancing mobile manipulation tasks in complex environments.

% \subsection{Embodied Dataset}
\vspace{0.5em}
\noindent\textbf{Embodied Dataset}.
In recent years, large-scale embodied datasets have become essential for advancing both navigation and manipulation tasks in robotics. 
On navigation, several datasets have been proposed to tackle challenges in object-goal navigation~\cite{ramakrishnan2021habitat,xia2018gibson,chang2017matterport3d,khanna2024habitat}, rearrangement~\cite{weihs2021visual,mirakhor2024task,batra2020rearrangement}, vision-language navigation~\cite{anderson2018vision,qi2020reverie,ku2020room,krantz_eccv20}, and question answering~\cite{das2018embodied,wijmans2019embodied,das2020probing,yu2019multi,Zhang2024QuestionguidedKG}. 
In contrast, manipulation-focused works leverage large-scale proprioceptive and visual data (e.g., RGB images or 3D point clouds) to simulate real-world dexterous hand operations~\cite{xu2023unidexgrasp,wan2023unidexgrasp++,wang2023dexgraspnet,Li2024SimultaneousDA,liu2024stat}, 
while other datasets provide object affordance labels to facilitate interaction learning~\cite{gao2024learning,brahmbhatt2020contactpose,corona2020ganhand,fan2023arctic,jian2023affordpose,taheri2020grab,yang2022oakink,zhan2024oakink2}. 
In parallel to simulation-based approaches, several real-world datasets have emerged to address the gap between simulated and physical environments. Some efforts like RoboTurk~\cite{Mandlekar2018ROBOTURKAC}, MIME~\cite{Sharma2018MultipleIM}, RoboNet~\cite{Dasari2019RoboNetLM} provide valuable demonstrations of physical object interactions, and FastUMI~\cite{zhaxizhuoma2025fastumiscalablehardwareindependentuniversal} propose a scalable hardware-independent robotic manipulation data collection system, while datasets such as KITTI~\cite{Geiger2013VisionMR}, nuScenes~\cite{Caesar2019nuScenesAM}, and Waymo Open~\cite{Sun2019ScalabilityIP} have significantly advanced autonomous navigation. More recently, research has increasingly focused on integrating manipulation capabilities with navigation for everyday tasks. Recent frameworks like BEHAVIOR Robot Suite~\cite{Jiang2025BEHAVIORRS} enable whole-body manipulation with bimanual coordination in household environments, while AgiBot World Colosseo~\cite{AgiBotWorldContributors2025AgiBotWC} offers over one million trajectories across 217 tasks. Despite these advances, real-world datasets still face challenges in human supervision and scalability~\cite{Pinto2015SupersizingSL,Levine2016LearningHC}. 
Moreover, some approaches use vision-language models or procedural methods to autonomously generate scalable language annotations in simulated environments~\cite{ha2023scaling,ehsani2024spoc,ahn2024autort,wang2023robogen,grauman2022ego4d,wang2023gen,duan2024manipulate,liu2024coherent,Tang2024Any2PointEA,Tang2023PointPEFTPF}. 
Despite these advances, navigation datasets often offer rich spatial information yet lack guidance for optimal positioning during subsequent manipulation, whereas manipulation datasets\textemdash despite providing valuable interaction data\textemdash do not fully capture the complexities of achieving optimal grasping positions via navigation.
\section{Conclusion}
\label{sec:conclusion}

Our work addresses the ``last mile'' challenge in mobile manipulation by introducing \ours, the first large-scale dataset (127k+ episodes across 569 kitchen scenes) featuring comprehensive, high-quality ground truth affordance maps to guide optimal positioning for manipulation tasks. We resolve the navigation-manipulation disconnect through automated cross-platform data collection and a unified framework compatible with diverse robotic systems, ensuring generalizability. Experimental results demonstrate that our baseline model \ourmodel achieves robust affordance prediction, validating the approach's efficacy. This dataset and methodology not only advance integrated navigation-manipulation systems for real-world deployment but also provide essential infrastructure for scalable and adaptive robotic learning in dynamic, cluttered environments.

% \clearpage
{
    \small
    \bibliographystyle{ieeenat_fullname}
    \bibliography{main}
}
\clearpage
\maketitlesupplementary

\setcounter{section}{0}
    This supplementary material extends our main study by providing additional details and data to improve the reproducibility of our \ours method. It includes further evaluations and a range of qualitative results for \ourmodel, which reinforce the conclusions drawn in the primary paper. Additionally, we offer some affordance collection videos in the accompanying zip file. \\
    \noindent $\triangleright$ \textbf{\cref{sec:dataset_sup}}: Describes the hierarchical structure of the dataset, including scenes, configurations, and episodes, with detailed information on the generation process, target objects, and simulation settings.\\
    \noindent $\triangleright$ \textbf{\cref{sec:implementation_sup}}: Provides an in-depth explanation of the evaluation metrics used, training configurations, and the baseline models compared in our study.\\
    \noindent $\triangleright$ \textbf{\cref{sec:add_exp}}: Presents additional visualizations of predictions, further performance evaluations, ablation results regarding the weight of the MSE loss, some data collection videos, and real-world demo video.\\
    \noindent $\triangleright$ \textbf{\cref{sec:limitation_future}}: Discusses the limitations of our work and explores prospects for future research.\\

\begin{figure*}[ht]
    \centering
    \includegraphics[width=0.98\textwidth]{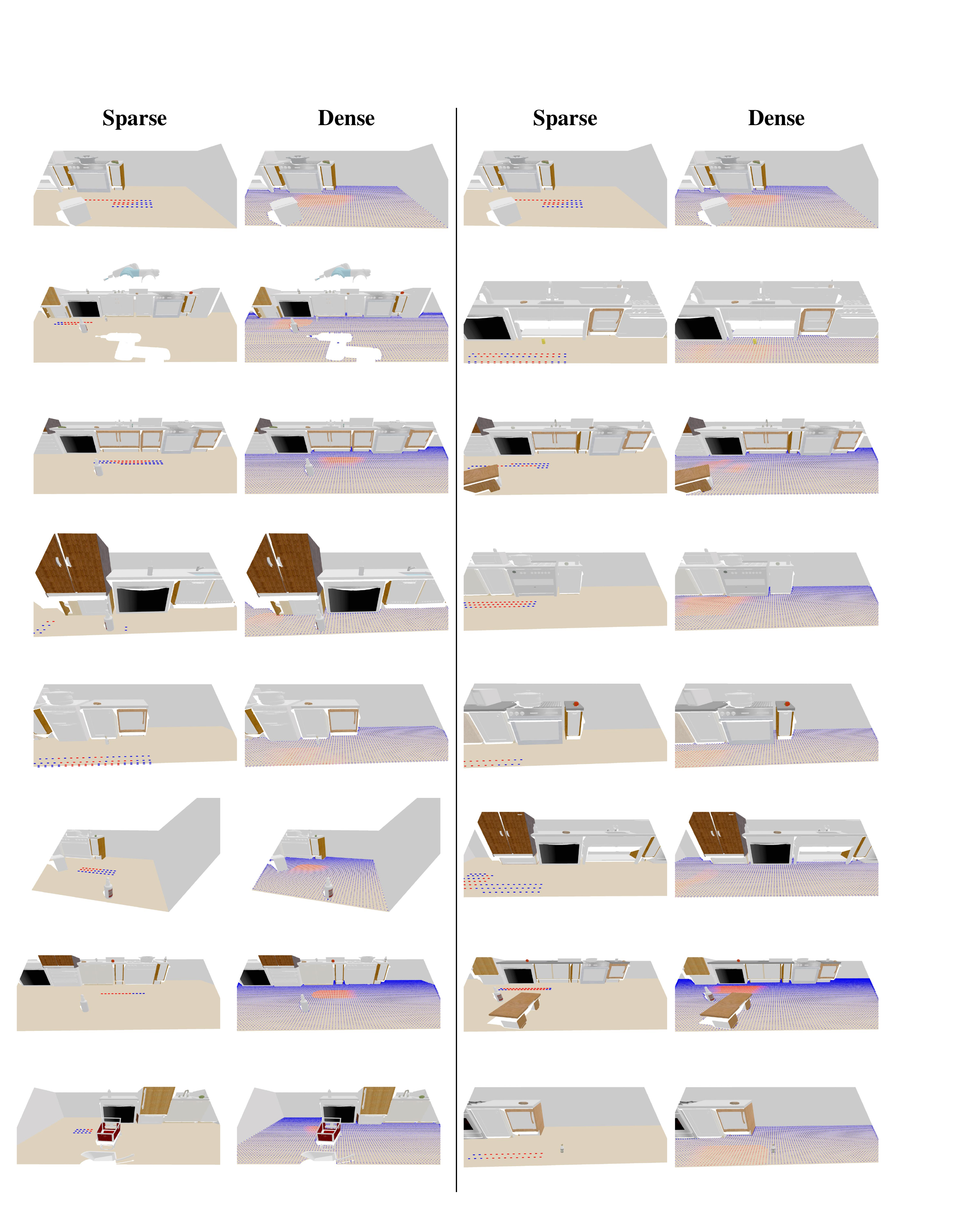}
    \caption{\textbf{Visualization of navigation affordance maps in \ours.} Comparison between sparse affordance values collected through discrete robot-object interactions (left) and their corresponding dense maps generated via Gaussian-weighted k-nearest interpolation (right).}
    \label{fig:more_vis_affgts}
\end{figure*}

\section{Dataset}
\label{sec:dataset_sup}

\subsection{Dataset Composition and Splits}

Our \ours is hierarchically organized into three levels: scenes, configurations, and episodes. Here, we provide a detailed description of each level.

A scene consists of randomly generated base furniture and layout, where certain articulated objects in the base furniture (\eg, microwaves, oven counters) serve as potential target objects for robotic arm manipulation. To ensure scene diversity, we randomly sample furniture categories, arrangement sequences, and specific instances within categories during scene generation. The statistics of target object assets employed in \ours are summarized in~\cref{tab:data_split_supp}.

Within each scene, we randomly place a varying number ($1$-$3$) of rigid objects, which, together with the articulated objects, constitute the set of target objects. To increase scene complexity, we position obstacles around these target objects. Each unique combination of target objects and obstacles forms a configuration of the scene.

To facilitate first-person view data collection, we sample $10$ views for each configuration using a camera mounted on the robotic arm. Each view generates one episode, and the view selection follows two principles: (1) views are randomly initialized around the target, and (2) views must encompass both the target object and the surrounding floor area.

In total, our \ours comprises $569$ scenes, $14,155$ configurations, and $127,343$ episodes, representing a comprehensive collection of mobile manipulation scenarios.

\begin{table}[t]
  \centering
  \resizebox{\linewidth}{!}{
    \begin{tabular}{c|cccccc}
      \toprule
      Rigid-Cats & All & Bottle & Pot & Fruit & Medicine Bottle & Vegetable \\
      & &
      \includegraphics[width=0.07\linewidth]{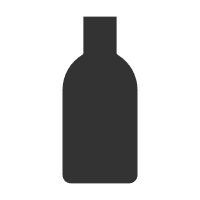} &
      \includegraphics[width=0.07\linewidth]{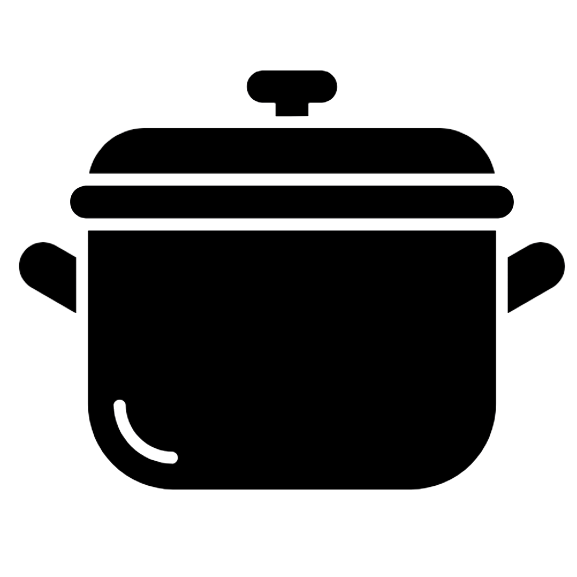} &
      \includegraphics[width=0.07\linewidth]{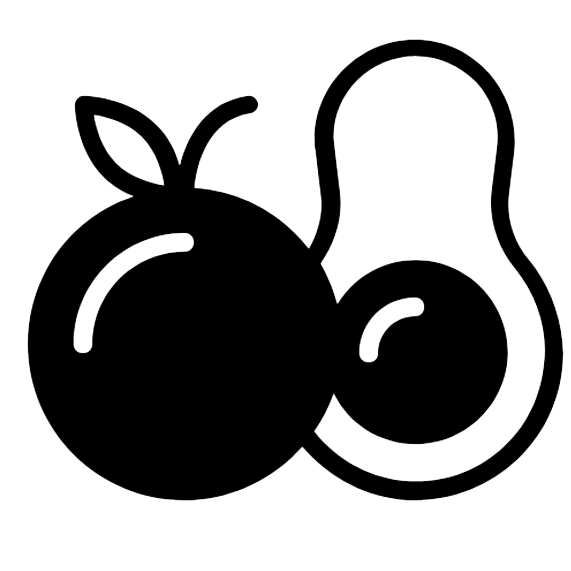} &
      \includegraphics[width=0.07\linewidth]{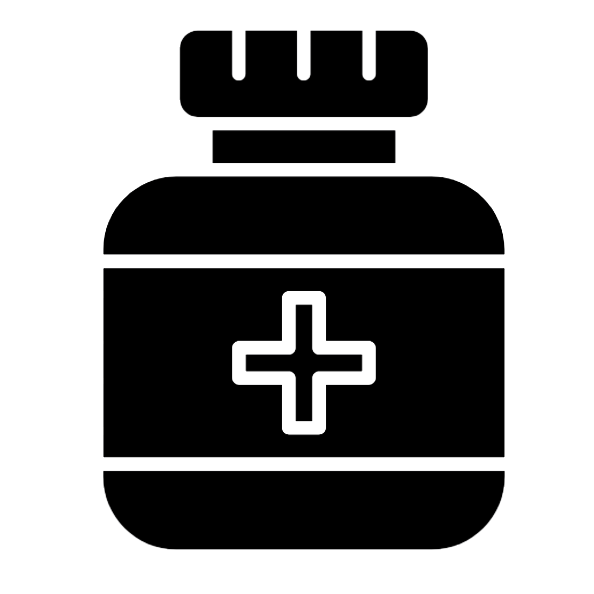} &
      \includegraphics[width=0.07\linewidth]{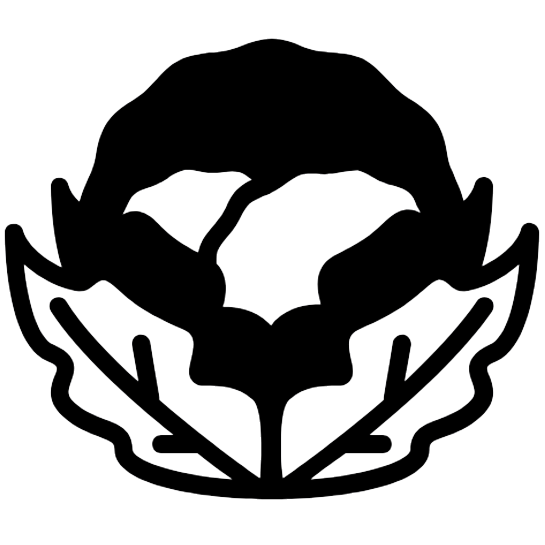} \\
      \midrule
      Rigid-Num & 65 & 6 & 7 & 11 & 8 & 33 \\
      \midrule
      Articulated-Cats & All & Faucet & Microwave & Cabinet & Dishwasher & Owen Counter \\
      & &
      \includegraphics[width=0.07\linewidth]{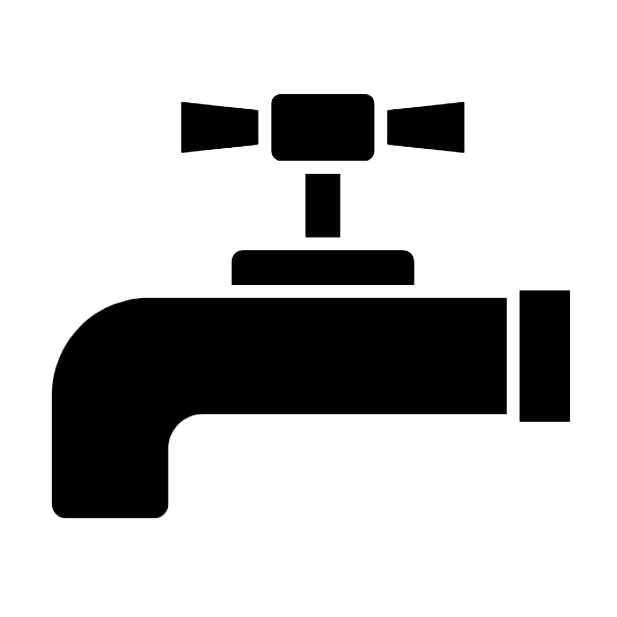} &
      \includegraphics[width=0.07\linewidth]{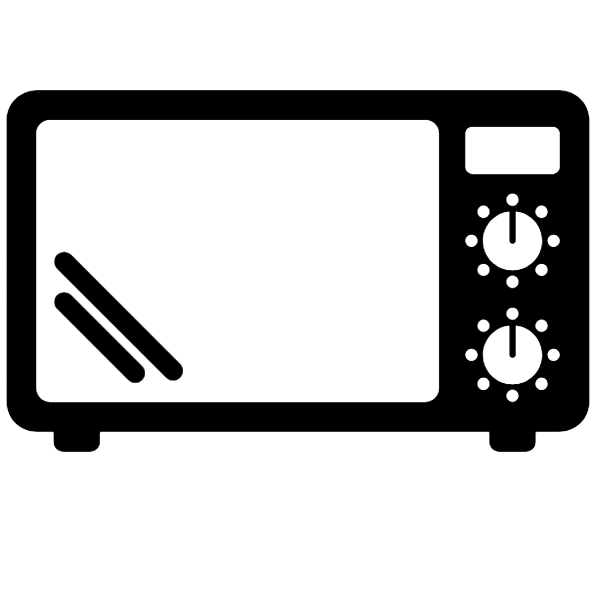} &
      \includegraphics[width=0.07\linewidth]{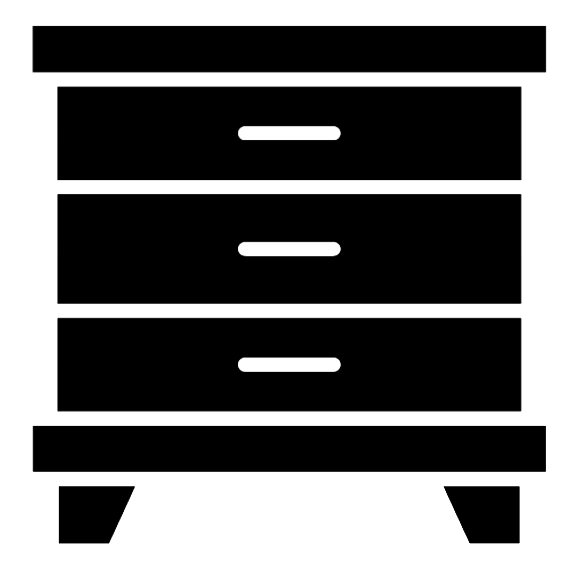} &
      \includegraphics[width=0.07\linewidth]{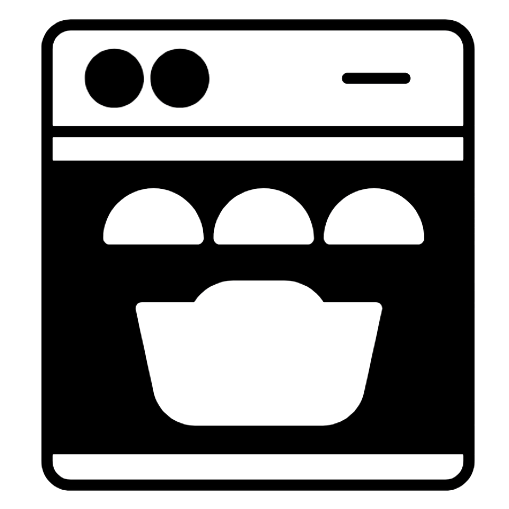} &
      \includegraphics[width=0.07\linewidth]{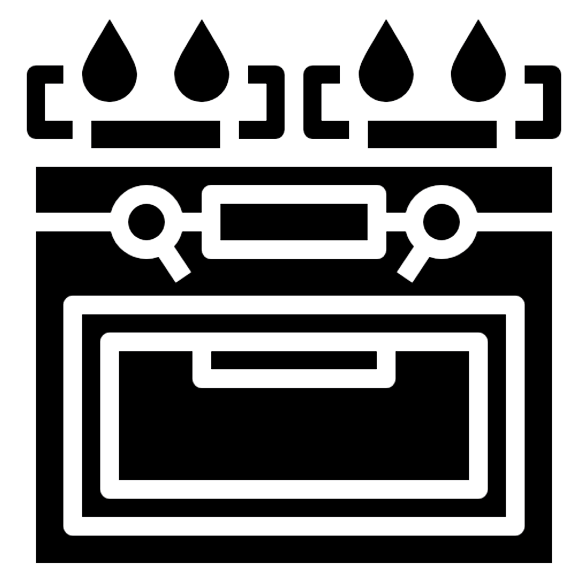} \\
      \midrule
      Articulated-Num & 48 & 11 & 7 & 20 & 1 & 9 \\
      \midrule
      Obstacle-Cats & All & Chair & Trolley & Bin & Table & Cart \\
      & &
      \includegraphics[width=0.07\linewidth]{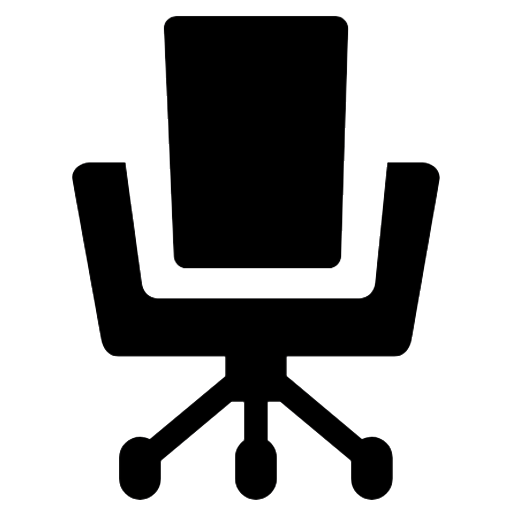} &
      \includegraphics[width=0.07\linewidth]{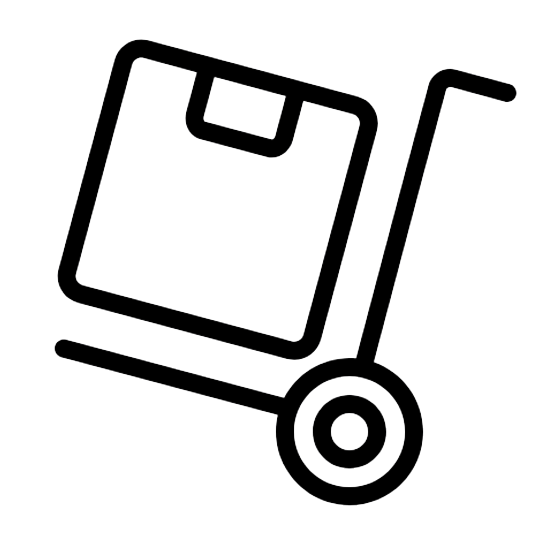} &
      \includegraphics[width=0.07\linewidth]{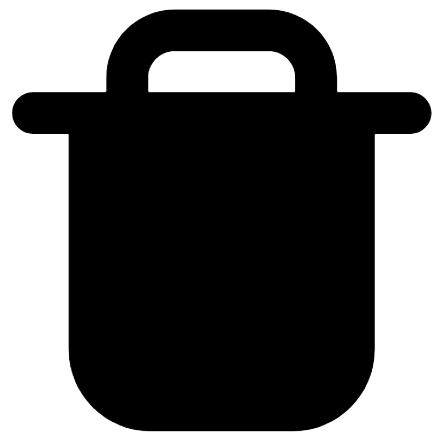} &
      \includegraphics[width=0.07\linewidth]{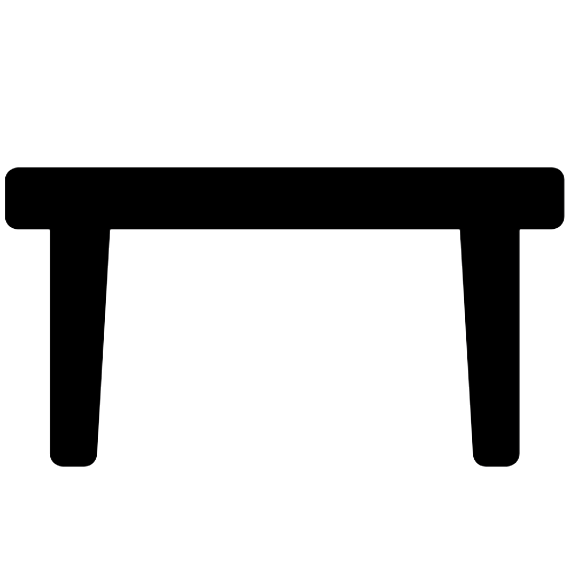} &
      \includegraphics[width=0.07\linewidth]{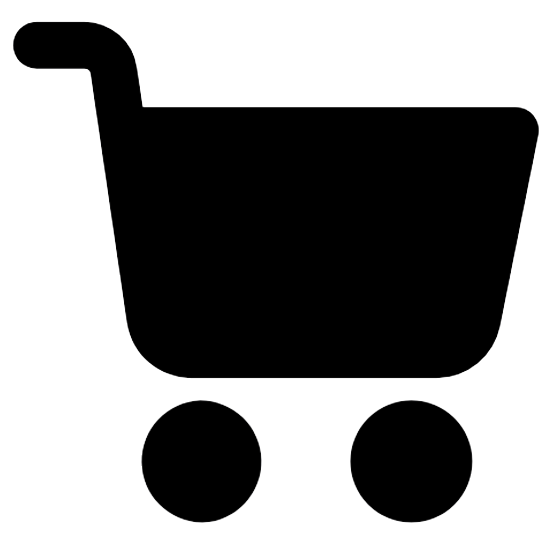} \\
      \midrule
      Obstacle-Num & 24 & 1 & 8 & 1 & 10 & 4\\
      \bottomrule
    \end{tabular}%
  }
  \caption{\textbf{Statistics of target object assets employed in \ours.} Distribution of rigid, articulated, and obstacle objects across different categories, showing the number of instances per category used in our dataset configurations.}

  \label{tab:data_split_supp}
\end{table}

\subsection{Details on Simulation}

We build our \ours based on the BestMan~\cite{Yang2024BestManAM} simulation environment, maintaining consistent simulation parameters across all scenes and interaction trials. The detailed configuration of our simulation setup is specified below:

\begin{itemize}
    \item \textbf{RGBD rendering.} We render RGB images and depth maps using the BestMan interface. For comprehensive first-person view sampling, we position the camera at varying locations relative to the target object. Specifically, the camera is placed either to the left or right with a lateral offset ranging from $0.0$ to $1.5$ meters, while the forward distance is sampled between $1.5$ and $3.8$ meters. The camera orientation is consistently directed toward the target object to ensure optimal coverage of both the target and the surrounding floor area. These sampling ranges were empirically determined to maximize viewpoint diversity while maintaining scene relevance.

    \item \textbf{3D point cloud.} We back-project the depth image into a 3D point cloud using the camera's intrinsic parameters. Subsequently, we filter out points with z-values below $0.02$ meters to obtain the floor point cloud.

    \item \textbf{Target objects sampling.} In each scene, we randomly position $1$-$3$ rigid objects in addition to the pre-existing articulated objects from scene generation. These objects collectively form our set of target objects, all of which are placed on kitchen countertops. To increase scene complexity, we randomly place $1$-$3$ obstacles within the semicircular region in front of each target object.

    \item \textbf{Interaction Trail.} To collect discrete navigation affordance values, we systematically sample robot positions within a semicircular region around the target object, with the radius set to the maximum reach of the robot arm. At each position, spaced at $10$ cm intervals along both x and y axes, the robot attempts to either grasp (for parallel grippers) or suction (for vacuum grippers) the target object.
\end{itemize}

\subsection{Additional Visualization Results}

We present additional visualization examples from \ours, including object assets, scene configurations, and affordance maps.

\subsubsection{Object Assets}
\begin{figure*}[ht]
    \centering
    \includegraphics[width=0.9\linewidth]{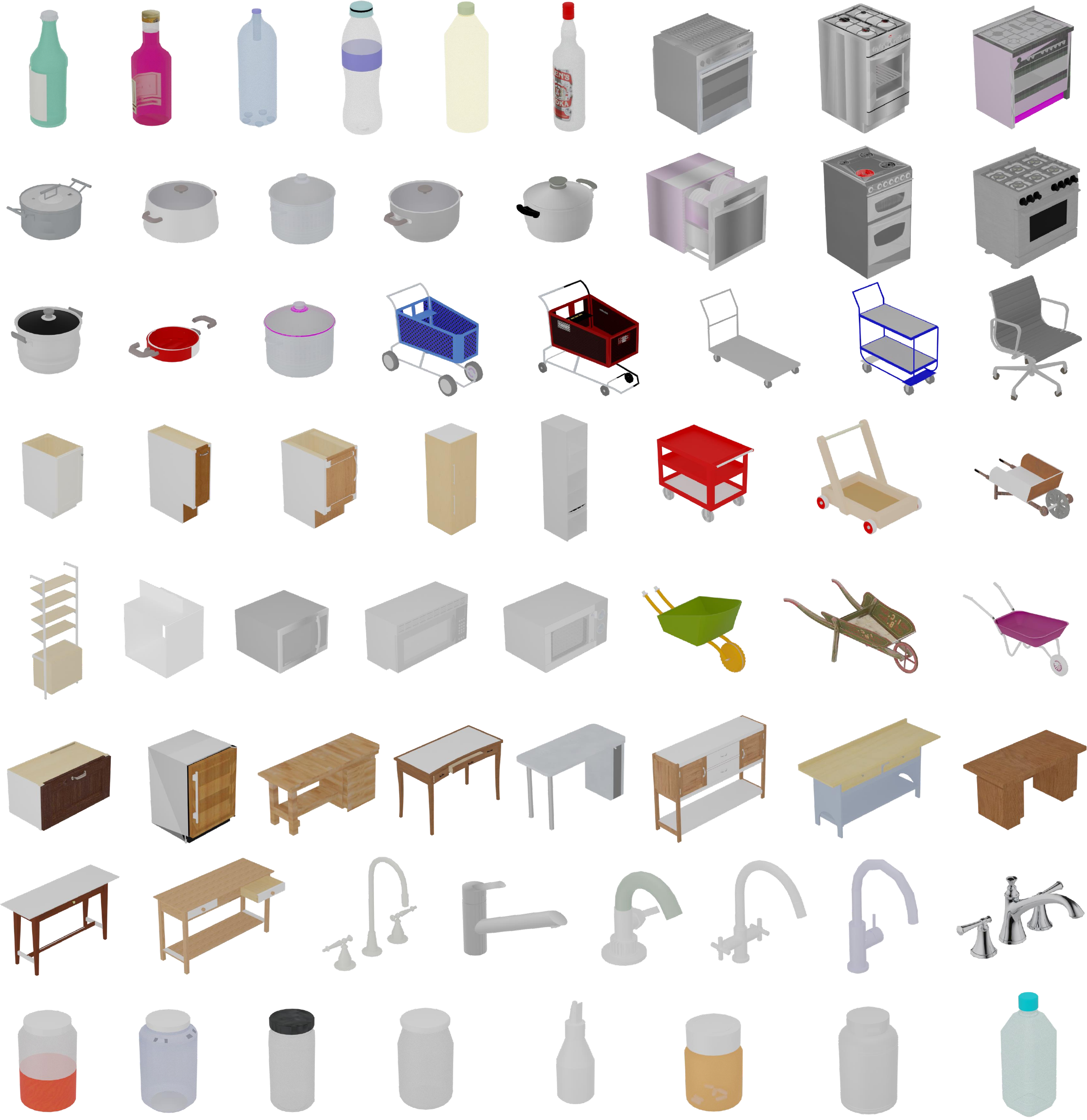}
    \caption{\textbf{Visualization of object assets in \ours.} The collection includes diverse categories of objects commonly found in household environments, ranging from kitchenware and appliances to furniture and daily necessities.}
    \label{fig:more_vis_object_assets}
\end{figure*}
We showcase a diverse set of object assets in~\cref{fig:more_vis_object_assets}. These assets serve as manipulation targets, environmental elements, or obstacles in our generated scenes.

\subsubsection{Scene Configurations}
\begin{figure*}[ht]
    \centering
    \includegraphics[width=0.8\linewidth]{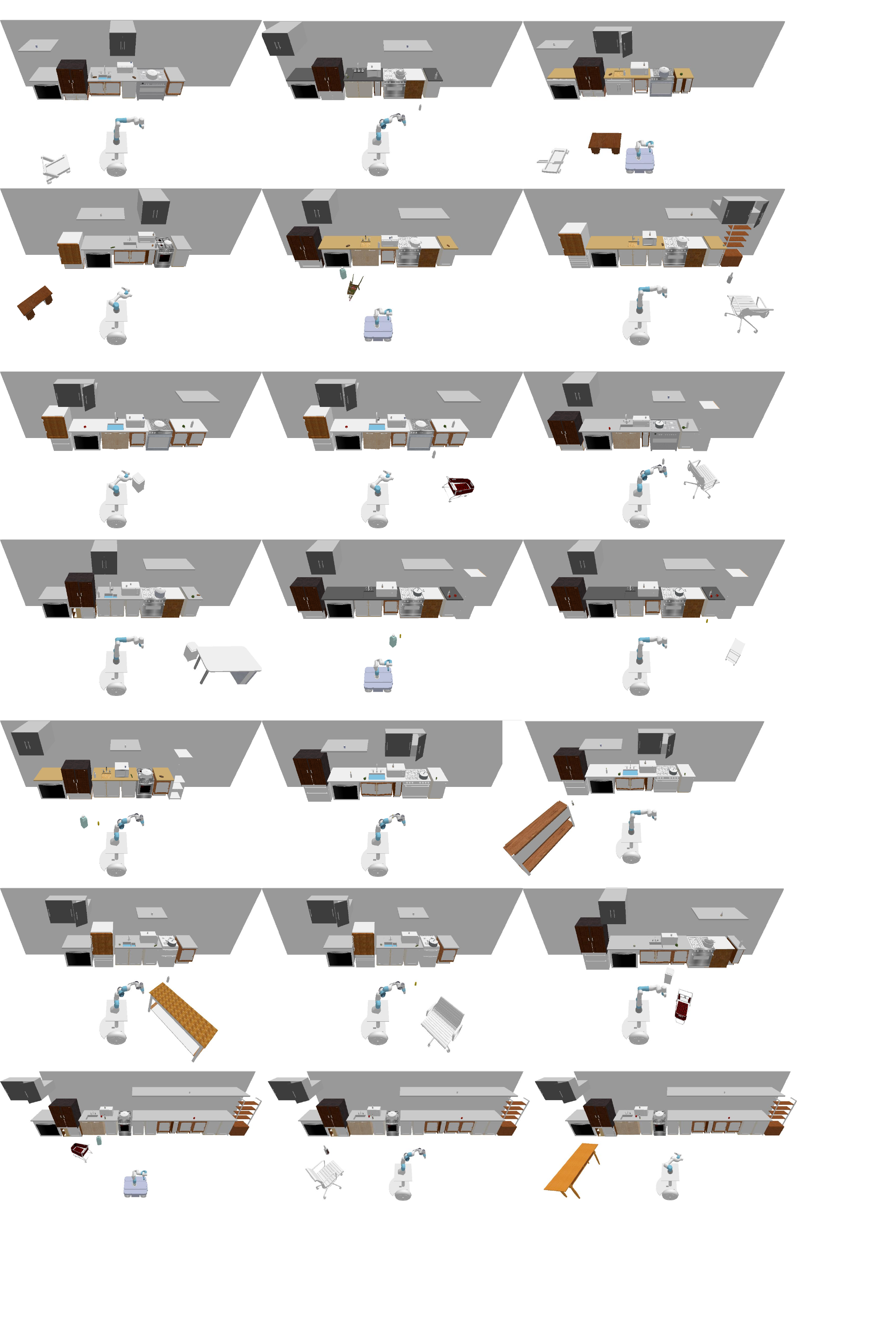}
    \caption{\textbf{Visualization of diverse scene configurations in \ours.} Each example showcases different arrangements of base furniture, layouts, target objects, and obstacles, demonstrating the variety of manipulation scenarios.}
    \label{fig:more_vis_scenes}
\end{figure*}

We illustrate a comprehensive set of scene configurations in~\cref{fig:more_vis_scenes}. These examples highlight the complexity and diversity of our generated environments, reflecting real-world manipulation scenarios.

\subsubsection{Affordance Map Examples}

We obtain sparse discrete navigation affordance values by systematically sampling robot positions and evaluating object interactions in the simulator. These sparse affordance values are then interpolated using a Gaussian-weighted k-nearest neighbor algorithm to generate continuous and dense navigation affordance maps. The comparison between sparse samples and interpolated dense maps is illustrated in~\cref{fig:more_vis_affgts}.

\section{Additional Implementation Details}
\label{sec:implementation_sup}

\subsection{Evaluation Metrics}
In this section, we provide a detailed explanation of the evaluation metrics used in our study. To comprehensively evaluate our method, we adopt five metrics: Root Mean Squared Error (\textbf{RMSE}), Logarithmic Mean Squared Error (\textbf{logMSE}), Pearson Correlation Coefficient (\textbf{PCC}), Cosine Similarity (\textbf{SIM}), and Continuous Intersection over Union (\textbf{cIoU}). Below, we describe each metric, its calculation formula, and its relevance to our task. 

\begin{itemize}
    \item \textbf{RMSE: }RMSE measures the numerical alignment between predicted and ground truth values. It penalizes larger errors through squared differences, as defined below: 
    \begin{align}
        \text{RMSE} &= \sqrt{\frac{1}{N} \sum_{i=1}^{N} (y_i - \hat{y}_i)^2}, 
    \end{align}
    where $y_i$ and $\hat{y}_i$ are ground truth and predicted values, respectively, and $N$ is the total number of elements. RMSE reintroduces the original scale of predictions, offering interpretability while highlighting large errors. In our study, RMSE assesses navigation affordance predictions by ensuring precise positioning and minimizing major errors in complex environments.

    \item \textbf{logMSE: }LogMSE evaluates the relative differences between predictions and ground truth, reducing the impact of large outliers. It is calculated as: 
    \begin{align} 
    \text{logMSE} &= \frac{1}{N} \sum_{i=1}^{N} \left(\log(1 + y_i) - \log(1 + \hat{y}_i)\right)^2, 
    \end{align}
    By focusing on proportional consistency, logMSE smooths out outliers and highlights relative accuracy. In this study, it evaluates the model’s ability to capture balanced affordance patterns across both low and high value regions.

    \item \textbf{PCC: }PCC quantifies the linear relationship between predicted and ground truth patterns, independent of their magnitudes: 
    \begin{align} 
    \text{PCC} &= \frac{\sum_{i=1}^{N} \left( y_i - \bar{y} \right)\left( \hat{y}_i - \bar{\hat{y}} \right)}{\sqrt{\sum_{i=1}^{N} \left( y_i - \bar{y} \right)^2 \sum_{i=1}^{N} \left( \hat{y}_i - \bar{\hat{y}} \right)^2}}, 
    \end{align}
    where $\bar{y}$ and $\bar{\hat{y}}$ are the means of $y$ and $\hat{y}$. PCC highlights pattern consistency, making it ideal for evaluating spatial distributions in affordance maps. A high PCC reflects accurate predictions of affordance trends, which is essential for precise navigation.

    \item \textbf{SIM: }SIM evaluates the alignment of relative spatial patterns between predictions and ground truth. It is calculated as: 
    \begin{align} 
    \text{SIM} &= \frac{1}{N} \sum_{i=1}^{N} \frac{ y_i \cdot \hat{y}_i }{ \| y_i \| \| \hat{y}_i \| }
    \end{align}

    SIM with higher values indicating better structural alignment. Unlike PCC, which evaluates overall pattern trends, SIM focuses on the overlap of affordance regions, making it particularly effective for assessing spatial alignment. In our study, SIM ensures that predicted affordance maps accurately capture the structural properties of ground truth.

\end{itemize}

\subsection{Training Details}
To ensure a fair comparison, we train our model and all baseline methods under consistent settings. The implementations for both our methods and the baselines are developed using PyTorch. All models are trained on a single NVIDIA A100 GPU with a batch size of 64 for 6 epochs, completing the entire process in approximately 8 hours. We utilize the Adam optimizer~\cite{kingma2017adammethodstochasticoptimization} with betas configured as 0.9 and 0.999. The learning rate is initialized at 8e-4 and follows a cosine decay schedule.

\subsection{Compared Baselines}

Since our work is the first to propose a benchmark for navigation affordance grounding, there are no existing methods that directly address this task. Therefore, we adapt several classical methods commonly used for feature extraction and 3D object detection on point clouds for comparison. Specifically, we include PointNet++\cite{qi2017pointnet++}, a foundational model for point cloud feature extraction, VoteNet\cite{Qi_2019_ICCV}, a pioneering method for 3D object proposal generation, and H3DNet~\cite{Zhang2020H3DNet3O}, which enhances object detection with hierarchical features. To ensure a fair and comprehensive comparison, we adapt the official implementations of these methods to our \ours dataset. Specifically, we reimplement their architectures and fine-tune them for the navigation affordance grounding task, conducting training and evaluation under identical experimental settings. This allows us to systematically assess their performance against our proposed \ourmodel. 

\noindent \textbf{PointNet++.} PointNet++ extends the original PointNet~\cite{qi2017pointnet} framework by introducing hierarchical feature learning for point cloud processing. This method divides the input point cloud into overlapping regions using a sampling and grouping strategy, applying PointNet locally to extract features, and aggregating them hierarchically. PointNet++ is widely used for tasks such as segmentation and classification in 3D point clouds. In our adaptation, the point cloud data is passed through an encoder to extract features, which are then decoded to predict navigation affordance. 

\noindent \textbf{VoteNet.} VoteNet introduces a deep Hough voting framework for 3D object detection in point clouds. The method employs a point-based network to generate votes for object centers, followed by an aggregation module that clusters votes to produce 3D object proposals. To adapt VoteNet for our benchmark, we removed the Vote Aggregation and Detection components originally used for bounding box regression, retaining the remaining modules to perform navigation affordance grounding. This adaptation ensures the network focuses on predicting affordance maps instead of object detection. 

\noindent \textbf{H3DNet.} H3DNet proposes a Hierarchical 3D Detection Network that improves 3D object detection by leveraging multi-level geometric features. The network integrates instance-level and part-level features using a coarse-to-fine detection pipeline and introduces novel feature aggregation modules to enhance geometric reasoning. Similar to VoteNet, in our adaptation, we removed the bounding box regression components while retaining the remaining modules to focus on navigation affordance grounding, enabling the network to predict affordance maps instead of object detection outcomes.

\section{Additional Experimental Results}
\label{sec:add_exp}

\begin{figure*}[t]
  \centering
  % \fbox{\rule{0pt}{2in} \rule{1.0\linewidth}{0pt}}
   \includegraphics[width=1.0\textwidth]{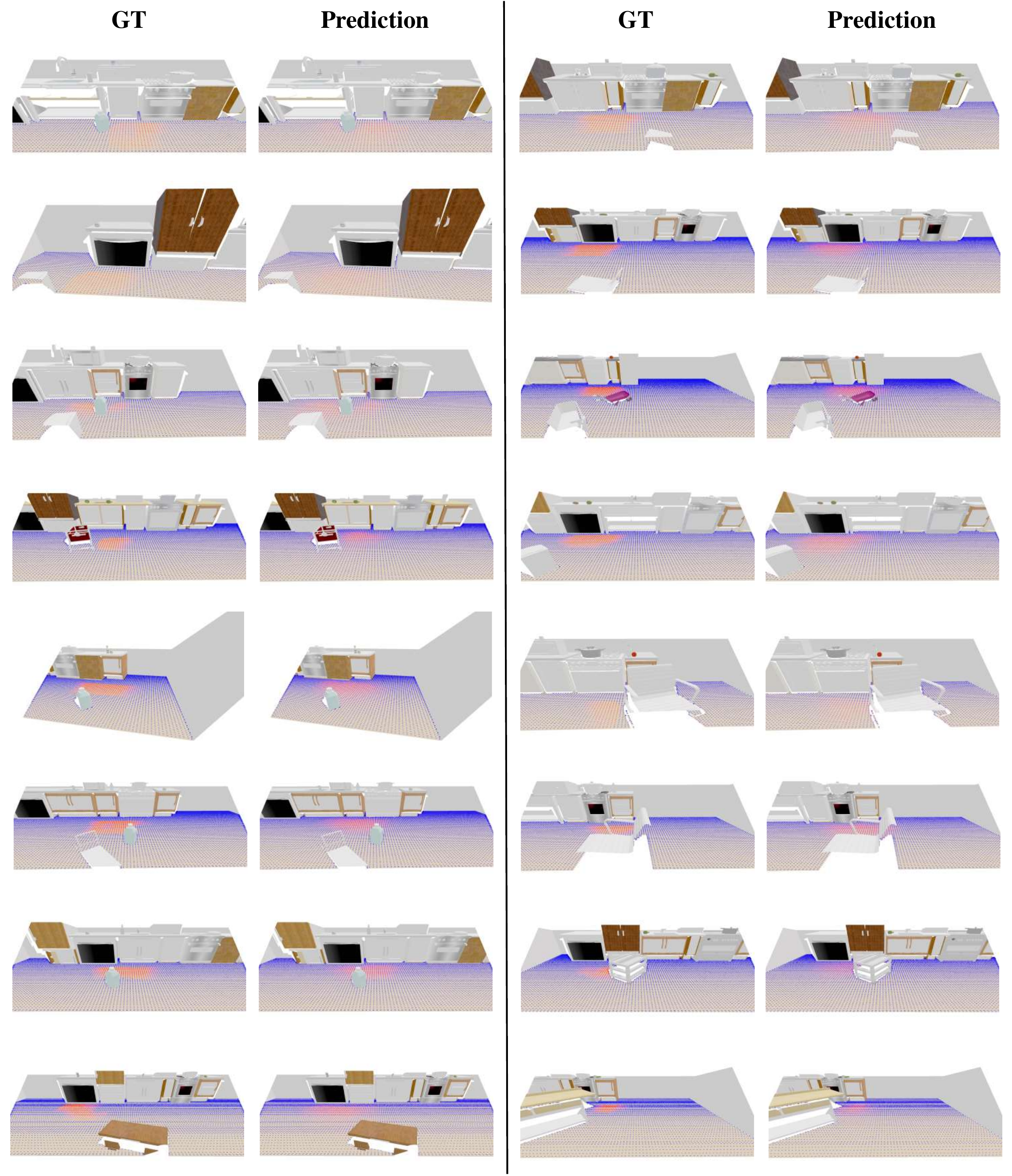}

   \caption{\textbf{Predicted vs. Ground Truth Navigation Affordance.} Comparison of the model's predicted navigation affordance (right columns) and the ground truth affordance (left columns) visualized within dense global point clouds. The visualizations illustrate the spatial alignment and consistency of the predictions with the reference data across different scenes. }
   \label{fig:pred_gt}
    % \vspace{-5pt}
\end{figure*}

\noindent \textbf{Visualization of Predictions.} As shown in ~\cref{fig:pred_gt}, we present the predicted navigation affordance results visualized within the dense global point cloud. Additionally, we provide a comparison with the ground truth affordance to highlight the model's performance and alignment with the reference data.

\noindent \textbf{Detailed Evaluation.} ~\cref{tab:detail} presents a comprehensive evaluation of various metrics within a single scene, offering an in-depth analysis of the model's behavior and performance. Each scene comprises multiple episodes, each representing distinct configurations and challenges. By evaluating metrics across these episodes, we gain a finer understanding of the model's ability to generalize under varying conditions. 

\begin{figure*}[t]
  \centering
  % \fbox{\rule{0pt}{2in} \rule{1.0\linewidth}{0pt}}
   \includegraphics[width=0.7\linewidth]{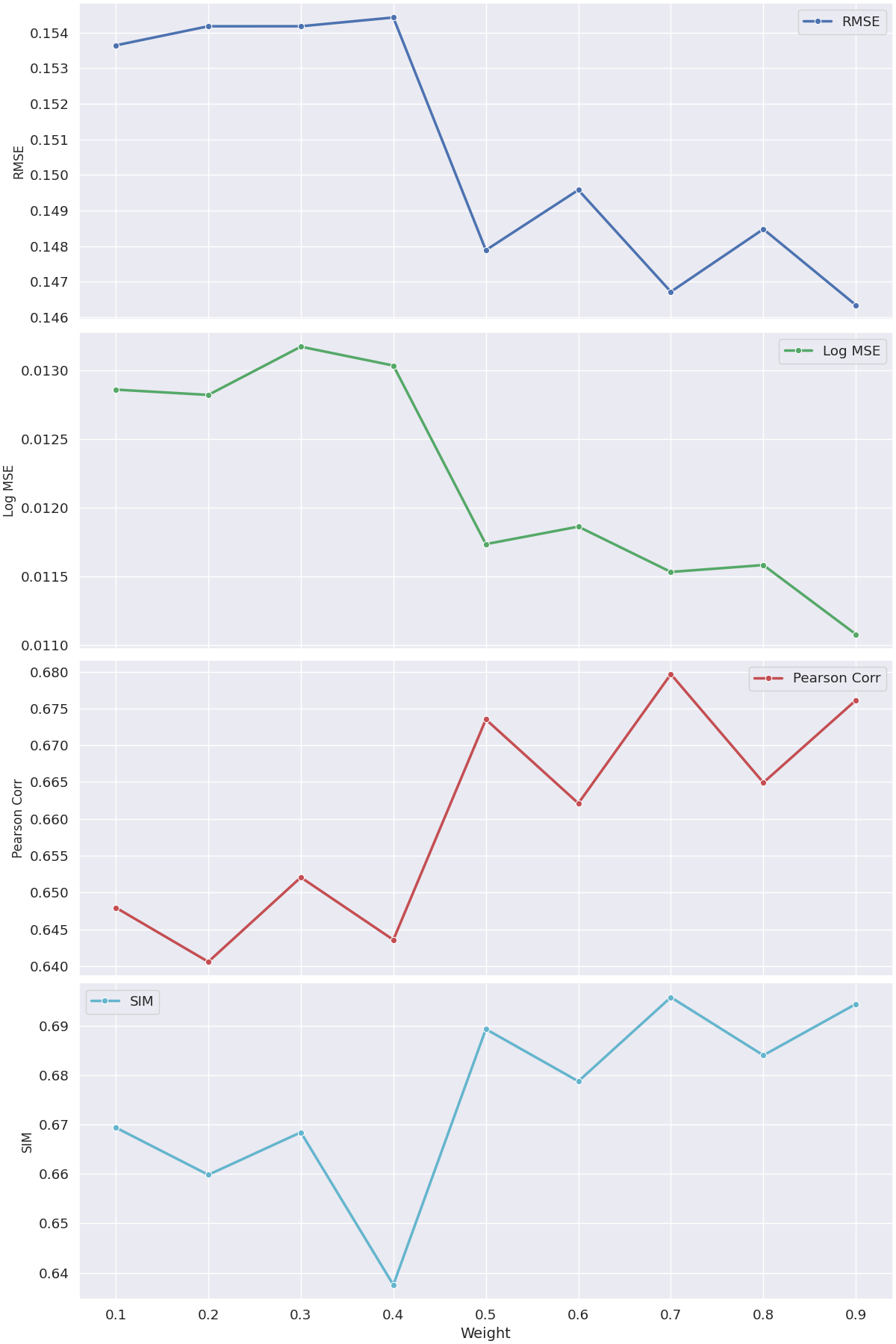}

   \caption{\textbf{Effect of Weight on MSE Loss and Evaluation Metrics.} Evaluation of the impact of different weight values in the Weighted MSE loss function on various metrics, including RMSE, LogMSE, PCC, and SIM.}
   \label{fig:metrics weight}
    % \vspace{-5pt}
\end{figure*}

\noindent \textbf{Impact of Weight Choices on Weight MSE Loss.}~\cref{fig:metrics weight} illustrates that when the weight value is too small, the model experiences underfitting because the influence of the loss function weight is insufficient, preventing the model from effectively learning high-quality navigation affordance grounding. Conversely, when the weight value is too large, the class imbalance issue described earlier persists, which also limits the model's performance. From the figure, it can be observed that when the weight value is set to 0.7, the Pearson Correlation Coefficient (PCC) reaches its peak. PCC measures the linear correlation between the predicted and ground truth values, effectively reflecting the model's ability to capture the distribution patterns of navigation affordance. A high PCC value indicates a stronger correlation between the predicted trends and the ground truth distribution, which is particularly critical for navigation tasks in complex environments.

\noindent \textbf{Real world Experiment.} Please see the real-world captured video demo in the zip file. 

\noindent \textbf{Affordance labeling visualization.} Please see the navigation affordance collection video in the zip file.

\section{Discussion on Limitations and Future Work}
\label{sec:limitation_future}

While our work makes significant progress in addressing the ``last mile'' navigation challenge, we acknowledge several limitations and identify promising directions for future research:

\begin{itemize}
\item \textbf{Scene Diversity:} Although \ours contains a large number of episodes, they are currently limited to kitchen environments. Future work should expand to other household scenarios such as living rooms, bedrooms, and bathrooms, which present different challenges and spatial configurations.

\item \textbf{Single-Task Focus:} The current approach focuses solely on reaching and grasping tasks. Future work should consider more complex manipulation sequences that require multiple positioning adjustments or different types of interactions (\eg, pushing, pulling, or sliding objects).
\end{itemize}

\paragraph{Future Directions}
\begin{itemize}
\item \textbf{Multi-Task Learning:} Future research could explore how navigation affordances vary across different manipulation tasks and develop models that can adapt their positioning strategies based on the intended manipulation action.

\item \textbf{Online Adaptation:} Developing methods for online adjustment of affordance predictions based on real-time feedback during task execution could enhance robustness in dynamic environments.

\item \textbf{Integration with LLMs:} While current LLM-based approaches have limitations, future work could explore hybrid approaches that combine our learned affordance models with LLM reasoning for more sophisticated task planning and execution.

\item \textbf{Uncertainty Estimation:} Incorporating uncertainty estimation in affordance predictions could help robots make more informed decisions about positioning and potentially trigger replanning when necessary.
\end{itemize}

These limitations and future directions present exciting opportunities for extending our work and further advancing the field of mobile manipulation.

\begin{table}[t]
    \centering
    \caption{\textbf {Evaluation results across individual scenes on \ours.} Performance metrics evaluated separately for each scene in the dataset. }
    \resizebox{1.0\linewidth}{!}{
        \begin{tabular}{ll|ccccc}
            \toprule
            &Scene ID  & \textbf{RMSE} $\downarrow$ & \textbf{logMSE} $\downarrow$ & \textbf{PCC} $\uparrow$ & \textbf{SIM} $\uparrow$  \\
            \midrule
            & 989172 & $0.283$ & $0.0439$ & $0.685$ & $0.702$ \\
            & 807952 & $0.269$ & $0.0402$ & $0.652$ & $0.667$ \\        
            & 502334 & $0.284$ & $0.0443$ & $0.716$ & $0.732$ \\
            & 306938 & $0.282$ & $0.0435$ & $0.737$ & $0.752$ \\
            & 443958 & $0.232$ & $0.0303$ & $0.615$ & $0.622$ \\
            & 66171 & $0.297$ & $0.0493$ & $0.595$ & $0.619$ \\        
            & 152285 & $0.242$ & $0.0332$ & $0.552$ & $0.578$ \\
            & 306168 & $0.223$ & $0.0282$ & $0.595$ & $0.609$ \\
            & 636942 & $0.283$ & $0.0445$ & $0.694$ & $0.713$ \\
            & 739657 & $0.264$ & $0.0392$ & $0.652$ & $0.672$ \\        
            & 143853 & $0.291$ & $0.0463$ & $0.742$ & $0.756$ \\
            & 739202 & $0.236$ & $0.0312$ & $0.561$ & $0.584$ \\
            & 583009 & $0.302$ & $0.0487$ & $0.702$ & $0.719$ \\
            & 451797 & $0.301$ & $0.0491$ & $0.728$ & $0.745$ \\ 
            & 116280 & $0.166$ & $0.0164$ & $0.305$ & $0.364$ \\
            & 274269 & $0.274$ & $0.0405$ & $0.829$ & $0.827$ \\
            & 485779 & $0.298$ & $0.0480$ & $0.730$ & $0.749$ \\
            & 772552 & $0.299$ & $0.0494$ & $0.683$ & $0.706$ \\
            & 359363 & $0.213$ & $0.0255$ & $0.566$ & $0.576$ \\
            & 264325 & $0.292$ & $0.0454$ & $0.593$ & $0.622$ \\        
            & 194561 & $0.276$ & $0.0417$ & $0.678$ & $0.695$ \\
            & 792629 & $0.292$ & $0.0462$ & $0.701$ & $0.715$ \\        
            & 69567 & $0.273$ & $0.0416$ & $0.722$ & $0.735$ \\
            & 783647 & $0.288$ & $0.0457$ & $0.644$ & $0.666$ \\
            & 721930 & $0.219$ & $0.0268$ & $0.500$ & $0.528$ \\        
            & 668061 & $0.266$ & $0.0385$ & $0.587$ & $0.616$ \\
            & 615543 & $0.275$ & $0.0408$ & $0.634$ & $0.648$ \\
            & 994972 & $0.285$ & $0.0436$ & $0.759$ & $0.754$ \\
            & 996121 & $0.266$ & $0.0389$ & $0.587$ & $0.604$ \\        
            & 67534 & $0.278$ & $0.0421$ & $0.633$ & $0.654$ \\
            & 142672 & $0.270$ & $0.0402$ & $0.700$ & $0.714$ \\
            & 501160 & $0.276$ & $0.0426$ & $0.692$ & $0.714$ \\
            & 487375 & $0.227$ & $0.0286$ & $0.519$ & $0.528$ \\        
            & 437964 & $0.294$ & $0.0465$ & $0.712$ & $0.729$ \\
            & 355986 & $0.299$ & $0.0485$ & $0.744$ & $0.764$ \\
            & 453020 & $0.285$ & $0.0449$ & $0.663$ & $0.686$ \\
            & 309033 & $0.296$ & $0.0465$ & $0.758$ & $0.762$ \\ 
            & 567413 & $0.225$ & $0.0284$ & $0.608$ & $0.619$ \\
            & 23304 & $0.248$ & $0.0340$ & $0.560$ & $0.586$ \\
            & 960190 & $0.247$ & $0.0336$ & $0.683$ & $0.699$ \\
            & 243997 & $0.288$ & $0.0454$ & $0.750$ & $0.762$ \\
            & 466622 & $0.265$ & $0.0389$ & $0.571$ & $0.592$ \\ 
            & 569661 & $0.278$ & $0.0424$ & $0.685$ & $0.705$ \\        
            & 403556 & $0.195$ & $0.0219$ & $0.481$ & $0.504$ \\
            & 297024 & $0.289$ & $0.0454$ & $0.724$ & $0.742$ \\        
            & 419493 & $0.219$ & $0.0278$ & $0.300$ & $0.369$ \\
            & 179882 & $0.213$ & $0.0255$ & $0.566$ & $0.625$ \\
            & 328786 & $0.256$ & $0.0370$ & $0.629$ & $0.643$ \\        
            & 475545 & $0.186$ & $0.0202$ & $0.573$ & $0.568$ \\
            & 773991 & $0.278$ & $0.0425$ & $0.665$ & $0.684$ \\        
            \bottomrule
            \end{tabular}
    }
    % \vspace{5pt}
    \label{tab:detail}
    \vspace{-8pt}
\end{table}

\end{document}